\documentclass{article}

\PassOptionsToPackage{numbers}{natbib}

\usepackage[eandd, preprint]{neurips_2026}

\usepackage[utf8]{inputenc} 
\usepackage[T1]{fontenc}    
\usepackage{hyperref}       
\usepackage{url}            
\usepackage{booktabs}       
\usepackage{amsfonts}       
\usepackage{nicefrac}       
\usepackage{microtype}      
\usepackage{graphicx}
\usepackage{enumitem}
\usepackage{fontawesome5}
\usepackage[dvipsnames,table,xcdraw]{xcolor}
\usepackage{amsmath}
\usepackage{multirow}
\usepackage{threeparttable}
\usepackage{todonotes}
\usepackage{wasysym}

\usepackage{amsthm}
\usepackage{thmtools}
\usepackage[most]{tcolorbox}

\declaretheoremstyle[
    spaceabove=6pt, spacebelow=6pt,
    headfont=\bfseries, headpunct={.}, headformat={\NAME\ \NUMBER},
    bodyfont=\normalfont,
    postheadspace=0.5em
]{promptstyle}

\declaretheorem[name=Prompt, style=promptstyle]{prompt}

\tcolorboxenvironment{prompt}{
    colback=gray!10!,
    colframe=gray!75!,
    fonttitle=\bfseries,
    title=Prompt,
    boxrule=0.5pt,
    sharp corners
}

\def\OURBENCH{AstroVLBench}

\newcommand{\arcsec}{\ensuremath{^{\prime\prime}}}

\usepackage{tabularx}

\setlist[itemize]{leftmargin=12pt}

\title{A systematic evaluation of vision-language models for observational astronomical reasoning tasks}

\author{%
  Wenke Ren \\
  Shanghai Astronomical Observatory\\
  Chinese Academy of Sciences\\
  Shanghai, China
  \And
  Hengxiao Guo \\
  Shanghai Astronomical Observatory\\
  Chinese Academy of Sciences\\
  Shanghai, China
  \And
  Wenwen Zuo \\
  Shanghai Astronomical Observatory\\
  Chinese Academy of Sciences\\
  Shanghai, China
  \And
  Xiaoman Zhang\thanks{Corresponding author.} \\
  School of Artificial Intelligence\\
  Shanghai Jiao Tong University\\
  Shanghai, China\\
  \texttt{xiaomanzhang.zxm@gmail.com}
}

\begin{document}

\maketitle

\begin{abstract}
Vision-language models (VLMs) are increasingly proposed as general-purpose tools for scientific data interpretation, yet their reliability on real astronomical observations across diverse modalities remains untested. We present \textbf{\OURBENCH}, a benchmark of over \textbf{4{,}100} expert-verified instances across five tasks spanning optical imaging, radio interferometry, multi-wavelength photometry, time-domain light curves, and optical spectroscopy. We systematically evaluate six state-of-the-art frontier models and find that performance is strongly modality-dependent and uncorrelated with general benchmark rankings. Mechanistic cross-model ablations reveal three concrete bottlenecks. First, prompt grounding exhibits marked model and task dependence: physical guidance yields more balanced classifications on optical imaging, whereas phenomenological prompts better navigate undersampled time-domain data. Second, presenting one-dimensional measurements as numerical tables rather than rendered plots consistently improves light-curve classification across all models by up to 23 percentage points, demonstrating that plot rendering can obscure physically relevant signal. Third, an expert audit of optical spectroscopy reveals a notable ``right-answer-wrong-reason'' phenomenon: across evaluated models, an average of 26\% of correct predictions are accompanied by physically inconsistent or invalid reasoning chains, showing that accuracy alone is insufficient for trustworthy scientific deployment. \OURBENCH{} provides the first systematic, multi-modal baselines for VLMs in observational astronomy and identifies the representation, grounding, and reasoning bottlenecks that must be addressed before VLMs can be trusted as autonomous scientific interpreters.
\end{abstract}


\section{Introduction}

\begin{figure}[!t]
    \centering
    \includegraphics[width=\textwidth]{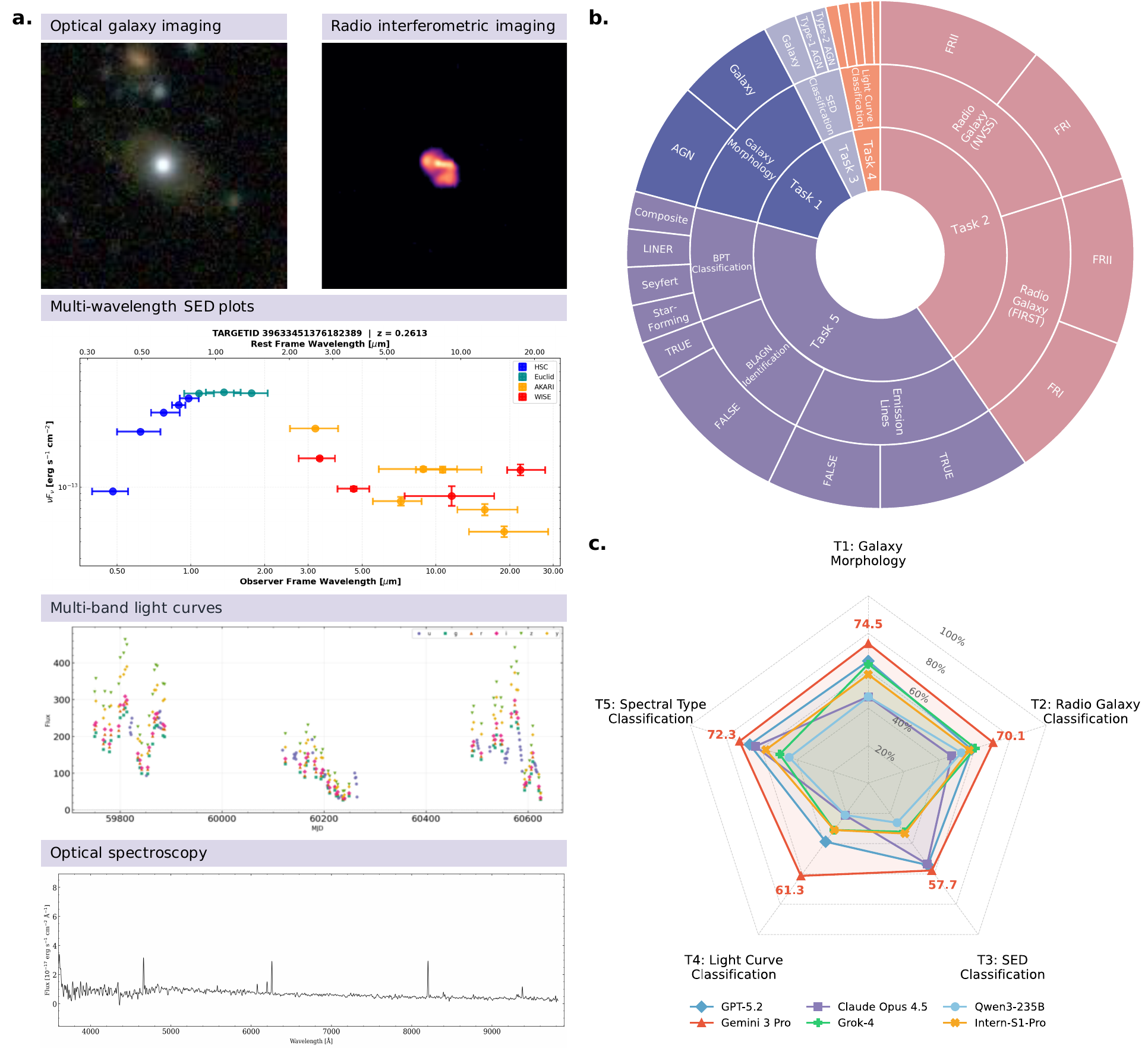}
    \caption{\textbf{Overview of \OURBENCH{} tasks and model performance.} (a)~Representative input examples for each of the five data modalities: optical galaxy imaging (Task~1), radio imaging (Task~2), multi-wavelength SED plots (Task~3), multi-band light curves (Task~4), and optical spectra (Task~5). Each task targets a distinct astronomical reasoning capability. (b)~Distribution of evaluation instances across all tasks and class labels. (c)~Radar plot comparing overall accuracy of the six frontier models across the five task categories under the guided baseline prompt condition.}
    \label{fig:overview}
\end{figure}

Observational astronomy is entering an era of unprecedented data scale and modality diversity. Large-area sky surveys across the electromagnetic spectrum, from the Dark Energy Spectroscopic Instrument (DESI)~\cite{DESICollaboration2025} to the Wide-field Infrared Survey Explorer (WISE)~\cite{Wright2010} and radio continuum surveys~\cite{Becker1995}, are delivering petabyte-scale datasets with unprecedented depth. Concurrently, time-domain facilities such as the Zwicky Transient Facility (ZTF)~\cite{Bellm2019}, the Vera C. Rubin Observatory~\cite{Ivezic2019}, and the forthcoming Square Kilometre Array Observatory (SKAO)~\cite{SKA2013} and Nancy Grace Roman Space Telescope~\cite{Roman2025} are accumulating billions of multi-wavelength light curves spanning decades of monitoring. This unprecedented scale and diversity of multi-modal data offer a pathway to a more comprehensive understanding of the physical processes governing galaxy formation and evolution. However, fully exploiting this scientific potential requires new paradigms: traditional equipment-specific pipelines lack the generalization needed to reason jointly across diverse modalities.

This fragmentation persists even within recent deep learning efforts. Early applications of deep learning in astronomy have predominantly focused on single modalities, functioning as isolated islands of expertise. Convolutional neural networks excel at galaxy morphology classification~\cite{Dieleman2015,Walmsley2022}, recurrent architectures are tailored for transient light-curve classification~\cite{Moller2020,SanchezSaez2021}, and specialized networks handle spectral energy distribution (SED) analysis~\cite{Boquien2019,Johnson2021}. While powerful within their specific domains, these task-specific models lack the capacity for cross-modal knowledge transfer, forcing astronomers to rely on a patchwork of specialized tools that each require separate training and maintenance. Furthermore, emerging multi-modal foundation models such as AION~\citep{Parker2025} and DeepDISC~\citep{Jiang2026} remain tied to specific telescope configurations and lack the interpretable reasoning chains required for scientific trust.

Vision-Language Models (VLMs) offer a promising path forward: a single general-purpose system that could replace the patchwork of task-specific pipelines. These systems jointly process images, tabular data, and natural language, and have demonstrated strong performance on tasks ranging from graduate-level physics problems~\cite{Rein2024} to mathematical discovery~\cite{RomeraParedes2024}. Early explorations in astronomy are encouraging, with models classifying optical transients~\cite{Stoppa2025}, fitting galaxy SEDs~\cite{Sun2024}, and identifying radio morphologies~\cite{Riggi2025,Drozdova2025}. Yet whether frontier VLMs can serve as general-purpose interpreters of real instrument data and where they fail when they cannot, remains largely unknown. Existing astronomy benchmarks fall short of answering this. AstroMLab~\cite{Ting2024,deHaan2025} probes text-based knowledge recall, while AstroMMBench~\cite{Shi2025} mixes textbook imagery with curated questions; neither tests reasoning over real, multi-modal observational data. Three concrete gaps result. First, no benchmark systematically tests frontier VLMs across multiple authentic observational modalities under controlled conditions. Second, when models fail, no existing evaluation disentangles whether the bottleneck is visual perception, missing domain knowledge, or flawed quantitative reasoning. Third, the relationship between accuracy and reasoning validity is largely unexplored, a critical omission, since a model that arrives at correct classifications via physically invalid logic poses serious risks for scientific deployment.

To address these gaps, we present \OURBENCH{}, a comprehensive benchmark for evaluating frontier VLMs on real observational data spanning five modalities fundamental to extragalactic and time-domain astronomy. Crucially, all five tasks are centered on Active Galactic Nuclei (AGN) classification, a fundamental galaxy-astronomy problem that naturally manifests across the entire electromagnetic spectrum. This unifying choice holds the underlying astrophysical concept approximately constant while varying only the observational modality, enabling a controlled comparison that disentangles a model's perceptual capabilities from its conceptual domain knowledge. The benchmark comprises over 4,100 expert-verified instances across five tasks: quasar-host discrimination from HSC optical imaging~\cite{Aihara2017} ($N=557$), Fanaroff-Riley classification on matched FIRST~\cite{Becker1995} and NVSS~\cite{Condon1998} radio maps from MiraBest~\cite{Porter2023} ($N=833$), AGN subtype classification from AKARI multi-wavelength photometry~\cite{Kim2020} ($N=168$), light-curve classification from PLAsTiCC simulated LSST photometry~\cite{malz2019photometric} ($N=142$), and hierarchical spectral interpretation on DESI DR1 spectra~\cite{DESICollaboration2025} ($N=1{,}600$ across three progressive subtasks). All ground-truth labels derive from spectroscopic confirmation or expert classification. All data, prompts, and evaluation suites are publicly available at \url{https://huggingface.co/datasets/XiaomanZhang/AstroVLBench}.

Using \OURBENCH{}, we systematically evaluate six state-of-the-art frontier models: four proprietary frontier models (GPT-5.2~\cite{GPT52}, Claude Opus 4.5~\cite{ClaudeOpus45}, Gemini~3~Pro~\cite{Gemini3Pro}, Grok-4~\cite{Grok4}) and two open-weight models (Qwen3-235B~\cite{Qwen3}, Intern-S1-Pro~\cite{InternS1Pro}) to ensure reproducibility. We complement headline accuracy with controlled cross-model ablations that probe the sources of success and failure. Our investigation is organized around five questions: How do zero-shot frontier VLMs compare to domain-specialized astronomical models? How does VLM competence vary across the five observational modalities? Does grounding model attention in physical mechanisms (\emph{why} a feature matters) outperform phenomenological guidance (\emph{what} to look for)? For one-dimensional measurements, do rendered plots or numerical tables better preserve physically relevant information? When models predict correctly, is the underlying reasoning physically valid?

Our main findings are summarized as follows:
\begin{itemize}[leftmargin=*, itemsep=1pt, topsep=2pt]
    \item \textbf{Zero-shot limitations versus domain specialists.} Frontier VLMs are competitive on spatial-morphology tasks but underperform specialized models overall. The gap widens sharply on tasks driven by precise numerical relationships, where zero-shot multi-modal reasoning cannot yet replace explicit feature extraction.
    \item \textbf{Modality-dependent competence.} Performance is strongly modality-dependent and decoupled from general benchmark rankings. While Gemini~3~Pro demonstrates the most consistent cross-modal competence overall, performance varies across observational regimes: top models converge to statistical parity on sparse SED classification, while individual models lead on specific subtasks.
    \item \textbf{Physical grounding and representation jointly shape robustness.} Prompt grounding exhibits model and task dependence: while explaining \emph{why} a feature matters promotes balanced classifications, phenomenological cues describing \emph{what} to look for remain necessary. For one-dimensional measurements, replacing rendered plots with numerical tables consistently improves accuracy across models by up to 23~percentage points.
    \item \textbf{The ``right-answer-wrong-reason'' phenomenon.} Accuracy and reasoning validity are frequently decoupled: an expert audit of optical spectroscopy reveals that on average 26\% of correct classifications are accompanied by physically inconsistent reasoning chains, demonstrating that reasoning validity is essential for trustworthy scientific deployment.
\end{itemize}

\section{Related Work}
\label{sec:related-work}

\paragraph{VLM benchmarks for science.}
Recent benchmarks have evaluated frontier vision-language models on scientific reasoning across diverse domains. GPQA~\cite{Rein2024} probes graduate-level knowledge in physics, chemistry, and biology through expert-written multiple-choice questions, while Humanity's Last Exam~\cite{Phan2025} pushes evaluation to the edge of expert knowledge. ScienceQA, MMMU, and related multi-disciplinary benchmarks~\cite{Phan2025,Rein2024} have shown that strong general capabilities do not always transfer to scientific subdomains. However, these benchmarks predominantly rely on textbook-style imagery and curated question-answer formats rather than authentic instrument data, leaving open how frontier models perform on raw observational measurements collected by real scientific facilities.

\paragraph{ML benchmarks in astronomy.}
Astronomy has a long history of single-modality classification benchmarks: Galaxy Zoo and its derivatives for optical morphology~\cite{Dieleman2015,Walmsley2022}, MiraBest~\cite{Porter2023} for radio galaxy morphology, PLAsTiCC~\cite{plasticc2019} for time-domain photometric classification, and DESI value-added catalogs~\cite{DESICollaboration2025} for spectral classification. These benchmarks have driven progress within their individual modalities through specialized convolutional, recurrent, and feature-engineered models~\cite{Moller2020,SanchezSaez2021,Boquien2019,Johnson2021,Boone2019}, but each operates on a single data type. The resulting performance numbers establish strong domain-specialist baselines but cannot be directly compared across modalities, and they leave unanswered whether a single general-purpose model can match these specialists across the observational landscape.

\paragraph{Foundation and language models for astronomy.}
A growing body of work explores foundation-model approaches in astronomy. AION~\cite{Parker2025} and DeepDISC~\cite{Jiang2026} build multi-modal representations tied to specific telescope configurations. AstroMLab~\cite{Ting2024,deHaan2025} introduces text-based multiple-choice evaluation of astronomical knowledge, and AstroMMBench~\cite{Shi2025} assembles 621 multimodal questions across six subfields, but both predominantly probe textual knowledge recall rather than reasoning over real instrument data. Application-side studies have used or fine-tuned VLMs for specific astronomical tasks, including transient classification~\cite{Stoppa2025}, multi-band SED fitting~\cite{Sun2024}, and radio morphology classification~\cite{Riggi2025,Drozdova2025}, while AI co-scientist frameworks~\cite{Gottweis2025} explore broader scientific reasoning workflows. \OURBENCH{} differs from this line of work in three respects: (i)~it evaluates frontier VLMs zero-shot on real observational data spanning five distinct modalities; (ii)~it controls for domain knowledge by centering all tasks on a unifying astrophysical concept (Active Galactic Nuclei classification); and (iii)~it goes beyond accuracy through prompt, representation, and reasoning-quality ablations that disentangle perceptual, conceptual, and representational sources of failure.

\section{Benchmark Construction} \label{sec:methods}

\subsection{Design and tasks}

\OURBENCH\ provides a systematic framework for assessing the astronomical reasoning proficiency of frontier VLMs. It encompasses five data modalities central to extragalactic and time-domain research (optical imaging, radio interferometric imaging, multi-wavelength photometric SEDs, time-domain photometric light curves, and optical spectroscopy) and over 4{,}100 evaluation instances curated from published surveys and catalogs. All ground-truth labels are derived from spectroscopic confirmation and expert identification, ensuring reproducibility. Centering all five tasks on Active Galactic Nuclei (AGN) classification controls for domain-knowledge confounding while still requiring distinct perceptual and reasoning capabilities at each modality.

\noindent \textbf{Task~1: QSO Host Galaxy Classification ($N=557$).}
Binary AGN-vs-Galaxy classification testing whether models can distinguish an unresolved nuclear point source (a Type-1 quasar dominated by a seeing-limited PSF) from the compact central emission of an inactive galaxy. We assemble 300 spectroscopically confirmed Type-1 AGN~\cite{Li2021} and 257 inactive galaxies~\cite{Simard2011} from HSC-SSP PDR3~\cite{Aihara2017,Aihara2022}, spanning a wide range of host-galaxy properties to avoid biasing toward any sub-population. Inputs are $g/r/i$ RGB composites generated with \texttt{trilogy}~\cite{Coe2012}, centered on the nucleus and spanning $10\arcsec\times10\arcsec$.

\noindent \textbf{Task~2: Radio Galaxy Morphology ($N=833$).}
Fanaroff-Riley Type~I (core-brightened) versus Type~II (edge-brightened) classification on MiraBest~\cite{Porter2023}, with 397 FRI and 436 FRII confidently classified sources. Each source has matched FIRST~\cite{Becker1995} (MiraBest\_F) and NVSS~\cite{Condon1998} (MiraBest\_N) maps, providing a direct measurement of how spatial resolution affects morphological classification.

\noindent \textbf{Task~3: SED Classification ($N=168$).}
Three-class classification (Type-1~AGN / Type-2~AGN / Galaxy) from multi-wavelength SEDs spanning optical to mid-infrared. We combine HSC ($g/r/i/z/y$), Euclid ($Y/J/H$), AKARI~NEP~\cite{Kim2020} (N2 to L24), and WISE (W1 to W4) photometry, cross-matched with the DESI~DR1 AGN/QSO Value-Added Catalog. The cohort size reflects the strict requirement of panchromatic coverage across all four facilities, as well as the intrinsic scarcity of luminous AGNs in the local universe ($\sim$5\% for Type-2 and $<1\%$ for Type-1~\cite{Asmus2020}). After photometric-completeness screening, we sample 44 Type-1~AGN, 40 Type-2~AGN identified by both NII-BPT~\cite{Baldwin1981} and SII-BPT~\cite{Veilleux1987} diagnostics, and 84 inactive galaxies. Inputs are $\nu f_\nu$ vs.\ wavelength plots (log-log) with rest-frame and observed-frame axes plus the spectroscopic redshift.

\noindent \textbf{Task~4: Light Curve Classification ($N=142$).}
Five-class classification across persistent variables and transients: AGN (stochastic), SNIa, TDE, RRL (short-period pulsators), Mira (long-period AGB pulsators). To avoid severe class imbalance from distorting evaluation metrics, we curate a balanced cohort of 30 sources per class from the PLAsTiCC simulated LSST dataset~\cite{plasticc2019}. The cohort size is bounded by the rarest classes in the challenge catalog (with well-sampled Mira variables limited to $N=22$, alongside astrophysically rare transients such as TDEs, of which only $\sim$200 confirmed optical events exist in observational literature~\cite{Langis2026}). Each curve presents flux-vs.-time across six LSST-like bands ($u/g/r/i/z/y$) with realistic survey cadences and noise.

\noindent \textbf{Task~5: Spectral Interpretation ($N=1{,}600$ across Q1/Q2/Q3).}
A hierarchical evaluation on DESI~DR1 optical spectra~\cite{DESICollaboration2025} that mirrors standard astronomical spectral analysis: line detection $\to$ profile recognition $\to$ ratio-based diagnostics. We curate 700 spectra in seven sub-groups (A, B, C1-C4, D) supporting three subtasks: \textbf{Q1} (700) Are both H$\alpha$ and H$\beta$ emission lines present? \textbf{Q2} (500) Is this a Broad-Line AGN? \textbf{Q3} (400) What is the BPT classification (Star-Forming, Composite, Seyfert, LINER)? Redshifts and line measurements come from \texttt{FastSpecFit}~\cite{Moustakas2023}; spectra are Gaussian-smoothed ($\sigma=5$ pixels).

\subsection{Evaluation design}

Beyond standard accuracy, we conduct controlled cross-model ablations and a quantitative reasoning audit across frontier models (Gemini~3~Pro, GPT-5.2, Claude Opus 4.5) to disentangle perceptual, conceptual, and representational sources of failure and evaluate cross-model generalizability.

\begin{itemize}[leftmargin=*, itemsep=1pt, topsep=2pt]
    \item \textbf{Perceptual versus conceptual guidance.} On Tasks~1 and~4, we compare three prompt conditions: \emph{unguided} (classification categories only), \emph{phenomenological} (visual features to look for, without physics), and \emph{physical} (astrophysical mechanisms producing those features).
    \item \textbf{Visual versus numerical input.} On Tasks~3 and~4, we compare informationally equivalent representations: a rendered plot versus a structured numerical table containing the exact flux/magnitude values, wavelengths or timestamps, and metadata. For Task~4 we also evaluate a shuffled-table control with randomly permuted timestamps.
    \item \textbf{Few-shot instruction.} On Tasks~2 and~5, we prepend labeled exemplar images as user-assistant turns. These modalities feature discrete, localized patterns with transferable cues, whereas Tasks~1, 3, and~4 involve continuous variations and wide dispersion ill-suited for sparse exemplars.
    \item \textbf{Model reasoning quality.} On Task~5, we evaluate reasoning quality through case studies and a systematic quantitative audit by an astronomy expert on sampled correct predictions, measuring the frequency of physically invalid justifications (excerpts and audit details in Appendix~\ref{sec:supp-reasoning}).
\end{itemize}

\subsection{Evaluation protocol}

All experiments use a two-message structure: a \emph{system message} containing the task description and classification guidelines (varying by condition), and a \emph{user message} containing a short instruction together with the image or text input. All prompts require structured JSON output \texttt{\{"answer": "", "reason": ""\}}, enabling automated extraction of predictions while capturing reasoning text for qualitative analysis. The user message is held constant across prompt conditions for each task; only the system message varies. Complete prompt templates are provided in Appendix~\ref{sec:supp-prompts}.

We evaluate six frontier VLMs spanning different organizations and architectural families: four proprietary frontier models (GPT-5.2~\cite{GPT52}, \texttt{gpt-5.2-2025-12-11}; Claude Opus 4.5~\cite{ClaudeOpus45}, \texttt{claude-opus-4-5-20251101}; Gemini~3~Pro~\cite{Gemini3Pro}, \texttt{gemini-3-pro-preview}; Grok-4~\cite{Grok4}, \texttt{grok-4}) and two open-weight models to ensure reproducibility (Qwen3-235B~\cite{Qwen3}, \texttt{qwen3-235b-a22b}; Intern-S1-Pro~\cite{InternS1Pro}, \texttt{intern-s1-pro}). All inputs are de-identified by removing celestial coordinates and target designations. All models are accessed via API with temperature 0 (greedy decoding) for deterministic, reproducible outputs. We report classification accuracy and balanced accuracy with 95\% class-stratified bootstrap confidence intervals (10{,}000 iterations, seed 42), and conduct paired randomization tests with Holm-Bonferroni correction for multiple comparisons.

\section{Benchmark Performance}

\begin{figure}[!t]
    \centering
    \includegraphics[width=\textwidth]{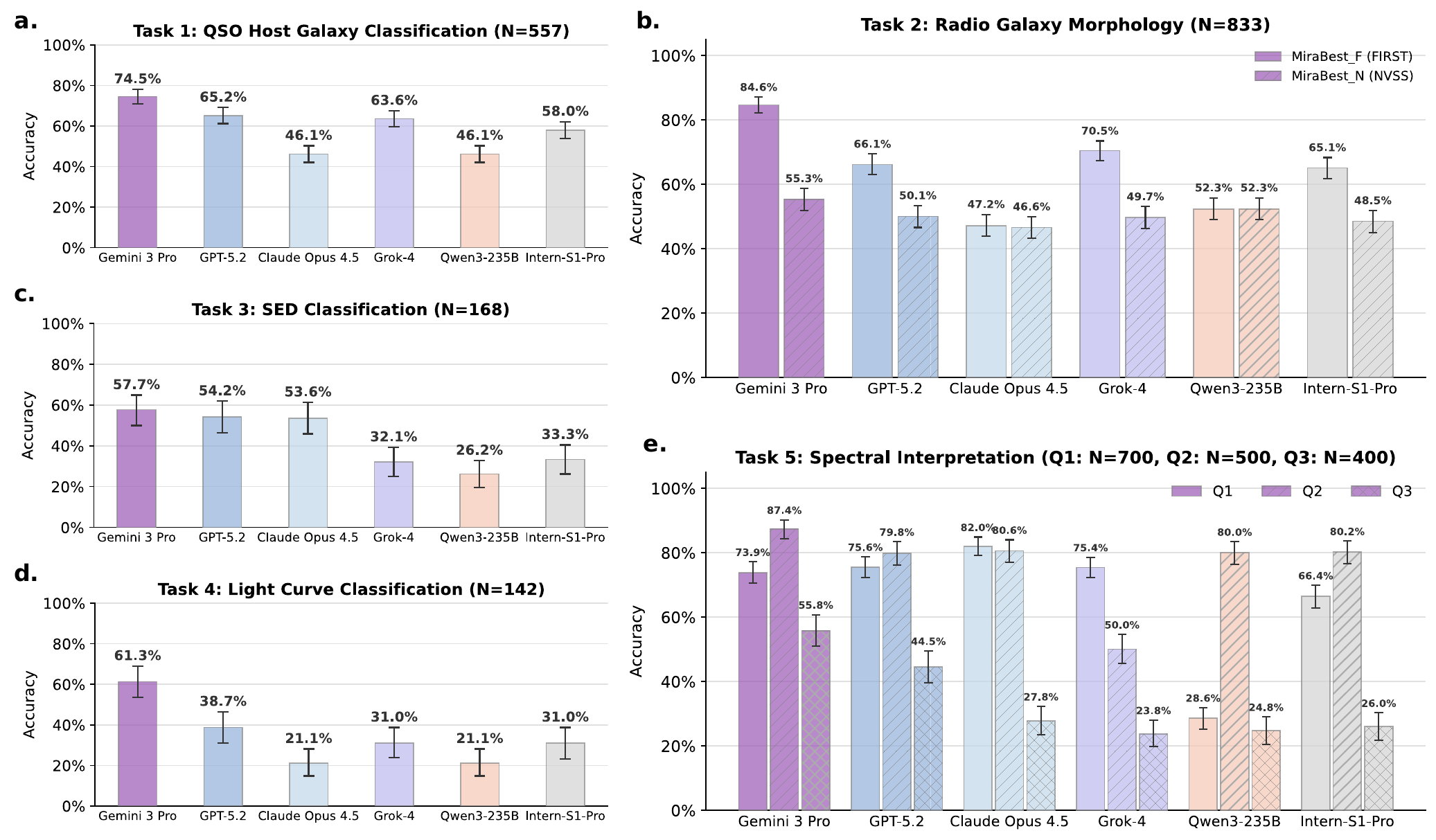}
    \caption{\textbf{Frontier VLM performance across five astronomical modalities.} Accuracy with 95\% bootstrap confidence intervals for six frontier models under the guided baseline prompt condition. (a)~Task~1: binary AGN/Galaxy classification ($N=557$). (b)~Task~2: Fanaroff-Riley classification on MiraBest\_F (FIRST, solid) and MiraBest\_N (NVSS, hatched), illustrating resolution dependence ($N=833$). (c)~Task~3: three-class SED classification ($N=168$). (d)~Task~4: five-class light curve classification ($N=142$). (e)~Task~5: hierarchical spectral interpretation Q1/Q2/Q3 ($N=700/500/400$), showing progressive difficulty. Models are ordered consistently across panels.}
    \label{fig:baseline}
\end{figure}

\subsection{Performance across observational modalities}
\label{sec:results-baseline}

Performance under the guided baseline prompt is summarized in Figure~\ref{fig:baseline}; complete tables and confusion matrices are in Appendix~\ref{sec:supp-tables} and~\ref{sec:supp-cm}. \textbf{Gemini~3~Pro is the most consistently capable model}, leading on optical imaging (Task~1, 74.5\% [0.709, 0.781]\footnote{All confidence intervals are bootstrap with 10{,}000 iterations.}), high-resolution radio morphology (Task~2 FIRST, 84.6\%), SED classification (Task~3, 57.7\%), light curve classification (Task~4, 61.3\%), and the harder spectral subtasks (Q2 87.4\%, Q3 55.8\%). Claude Opus 4.5 leads on Q1 (82.0\%) by maintaining 96\% specificity and 76\% recall; Gemini drops to 73.9\% on Q1 by being more conservative. Paired randomization tests with Holm-Bonferroni correction confirm that Gemini~3~Pro significantly outperforms other models on Tasks~1, 2 (FIRST), 4, and 5 (Q2, Q3) ($p < 0.05$), while Claude Opus 4.5 leads on Task~5 Q1 ($p < 0.05$). In contrast, models converge when observational constraints dominate: on multi-wavelength SEDs (Task~3), top models reach statistical parity within overlapping confidence intervals ($p > 0.05$), while lower spatial resolution on Task~2 NVSS similarly compresses performance across architectures (Appendix~\ref{sec:supp-stats}).

Three patterns recur across tasks. (i)~\emph{Modality dependence is large and uncorrelated with general benchmark rankings.} Models like Claude Opus and Qwen3 that score $\sim$85\% on text-based astronomy Q\&A~\cite{Ting2024} fail catastrophically on real observational tasks (e.g., 0\% AGN recall on Task~1; mode collapse on Tasks~2/4 and Q2/Q3 of Task~5). (ii)~\emph{Resolution and class-frequency drive failure modes.} On Task~2, NVSS's lower resolution drops every model below 56\%; on Task~3 binary AGN-vs-Galaxy, even Gemini reaches only 32\% AGN recall (and GPT-5.2 18\%) due to a strong conservative bias toward predicting Galaxy. (iii)~\emph{Difficulty grows monotonically with the reasoning chain.} The Q1$\to$Q2$\to$Q3 spectral hierarchy reveals a sharp transition: models can spot localized emission features (Q1) but mostly collapse on line-profile recognition (Q2) and ratio-based BPT diagnostics (Q3), where extracting and comparing quantitative amplitudes across wavelengths remains beyond robust capability for all but Gemini. Detailed per-task error analyses and confusion matrices are reported in Appendix~\ref{sec:supp-perf-details}.

\begin{figure}[!t]
    \centering
    \includegraphics[width=0.95\textwidth]{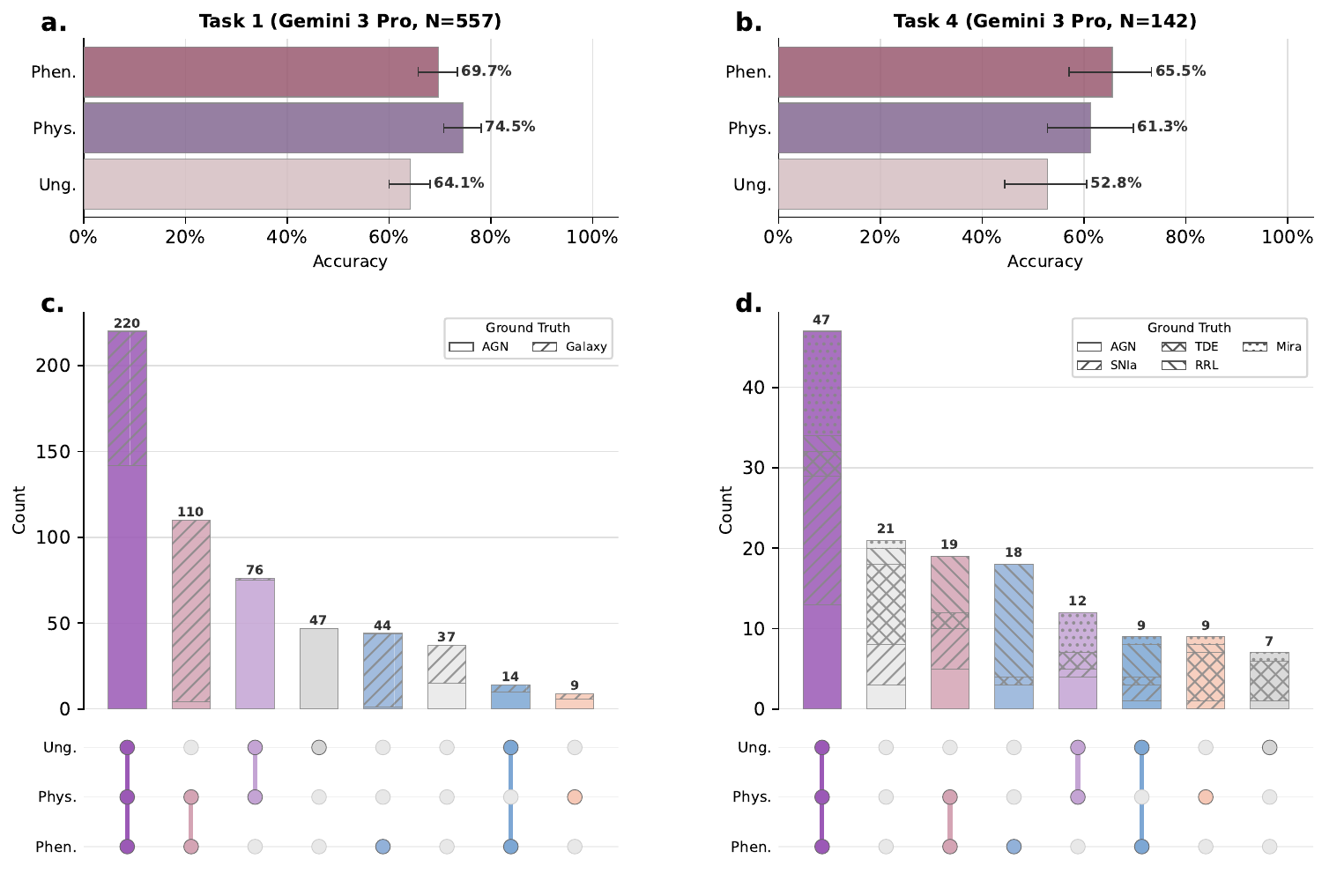}
    \caption{\textbf{Perceptual vs.\ conceptual guidance} (Gemini~3~Pro). \textbf{a,\,b.}~Accuracy under three prompt conditions (unguided, physical, phenomenological) with 95\% bootstrap CIs for Task~1 ($N=557$) and Task~4 ($N=142$). \textbf{c,\,d.}~UpSet plots showing case-level overlap across the three prompt conditions, bars stacked by ground-truth class.}
    \label{fig:prompt-guidance}
\end{figure}

\subsection{Perceptual versus conceptual guidance}
\label{sec:results-domain-knowledge}

We evaluate prompt guidance across frontier models (Gemini~3~Pro, GPT-5.2, Claude Opus 4.5) under unguided, physical, and phenomenological conditions (Figure~\ref{fig:prompt-guidance}, Table~\ref{tab:supp-prompt-ablation}, Appendix~\ref{sec:supp-cm-ablation}). Rather than a uniform hierarchy, prompt effects exhibit strong model and task dependence. On Task~1 (imaging), physical guidance establishes the most balanced decision boundaries for capable models (Gemini~3~Pro and GPT-5.2), whereas phenomenological descriptions bias models toward Galaxy predictions by emphasizing diffuse host features over compact nuclei. Claude Opus 4.5 hovers near random guessing (46\%--51\%), its nominal unguided lead reflecting a degenerate prior favoring galaxies.

On Task~4 (light curves), phenomenological guidance achieves the highest overall accuracy across \emph{all three models}, universally unblocking short-period pulsators (RR Lyrae). While physical prompts misattribute sparse-sampling artifacts to stochastic AGN variability (yielding $\le33\%$ RRL recall), articulating cadence-induced aliasing drives dramatic RRL recall surges across all architectures ($53\%$--$97\%$). Crucially, model capacity dictates representational stability: whereas Gemini~3~Pro maintains balanced discrimination across non-pulsating classes, lower-performing models exhibit heuristic over-correction (GPT-5.2 cannibalizes Miras and AGNs into RRL) or severe mode collapse (Claude confuses TDEs with SNIa). Thus, phenomenological prompts resolve observational sampling artifacts, but robust discrimination still requires strong baseline perceptual capability.

\begin{table}[!htbp]
    \centering
    \scriptsize
    \setlength{\tabcolsep}{3.5pt}
    \begin{minipage}[t]{0.47\linewidth}
        \centering
        \caption{\textbf{Visual vs.\ numerical input across models.} Accuracy (\%) on Tasks~3 (SED) and~4 (light curves) under plot and tabular formats.}
        \label{tab:representation}
        \begin{tabular}{llccc}
            \toprule
            Task & Representation & Gemini & GPT-5.2 & Claude \\
            \midrule
            \multirow{2}{*}{Task~3} & Image (plot)      & \textbf{57.7} & 38.1          & \textbf{35.1} \\
                                    & Table (numerical) & 51.8          & \textbf{57.7} & 25.6 \\
            \midrule
            \multirow{3}{*}{Task~4} & Image (plot)      & 61.3          & 35.9          & 30.3 \\
                                    & Table (ordered)   & \textbf{74.6} & \textbf{49.3} & \textbf{52.8} \\
                                    & Table (shuffled)  & 68.3          & 41.6          & 40.1 \\
            \bottomrule
        \end{tabular}
    \end{minipage}\hfill
    \begin{minipage}[t]{0.51\linewidth}
        \centering
        \caption{\textbf{Few-shot instruction across models.} Accuracy (\%) under zero-shot (0S) vs.\ few-shot visual exemplars (FS) on radio and spectral tasks.}
        \label{tab:fewshot}
        \begin{tabular}{lcccccc}
            \toprule
            \multirow{2}{*}{Subset} & \multicolumn{2}{c}{Gemini} & \multicolumn{2}{c}{GPT-5.2} & \multicolumn{2}{c}{Claude} \\
            \cmidrule(lr){2-3} \cmidrule(lr){4-5} \cmidrule(lr){6-7}
             & 0S & FS & 0S & FS & 0S & FS \\
            \midrule
            Task~2 FIRST & \textbf{84.6} & 83.1 & 58.3 & \textbf{59.7} & 39.1 & \textbf{51.7} \\
            Task~2 NVSS  & \textbf{55.3} & 53.3 & \textbf{52.1} & 50.7 & 44.9 & \textbf{47.9} \\
            Task~5 Q1    & 73.9 & \textbf{78.9} & 57.0 & \textbf{72.6} & 39.6 & \textbf{55.4} \\
            Task~5 Q2    & \textbf{87.4} & 86.4 & 81.4 & \textbf{84.8} & 77.6 & \textbf{81.6} \\
            \bottomrule
        \end{tabular}
    \end{minipage}
\end{table}

\subsection{Visual versus numerical input}
\label{sec:results-representation}

Tasks~3 and~4 render continuous measurements as plots; we evaluate representation effects by replacing images with structured numerical tables across frontier models (Table~\ref{tab:representation}). Representation efficacy diverges cleanly between static multi-wavelength profiles and dynamic time series. On Task~3 ($N=168$), visual plots outperform tables for Gemini (57.7\% vs.\ 51.8\%) and Claude (35.1\% vs.\ 25.6\%), consistent with SED classification depending on continuous spectral shapes and infrared slopes that visual curves naturally preserve, whereas GPT-5.2 benefits from explicit tabular fluxes (57.7\% vs.\ 38.1\%). Conversely, on Task~4 ($N=142$), tabular input yields consistent, substantial gains across all models: Gemini improves 61.3\%$\to$74.6\%, GPT-5.2 35.9\%$\to$49.3\%, and Claude 30.3\%$\to$52.8\% (+22.5~pp), directly bypassing cadence-induced visual ambiguities. Preserving temporal order is critical: the shuffled control degrades accuracy across models (dropping to 68.3\%, 41.6\%, and 40.1\%), though remaining above images. Thus, continuous SED profiles favor visual rendering, whereas time-domain measurements are vastly better preserved by numerical tables.

\subsection{Few-shot instruction}
\label{sec:results-fewshot}

We evaluate whether labeled visual exemplars improve classification on Tasks~2 and~5 across models (Table~\ref{tab:fewshot}). On Task~5 Q1, exemplars produce consistent, substantial improvements across models (Gemini: 73.9\%$\to$78.9\%, GPT-5.2: 57.0\%$\to$72.6\%, Claude: 39.6\%$\to$55.4\%), where Balmer emission features are localized, discrete, and stable. In contrast, on Task~2 radio morphology, few-shot demonstrations yield modest or negligible shifts (FIRST: Claude improves 39.1\%$\to$51.7\%, GPT-5.2 58.3\%$\to$59.7\%, Gemini shifts 84.6\%$\to$83.1\%; NVSS remains within 1--3\%), as few exemplars cannot cover the broad structural diversity of FR jets. Similarly, on Task~5 Q2, few-shot shifts are marginal (Gemini 87.4\%$\to$86.4\%, GPT-5.2 81.4\%$\to$84.8\%). Visual exemplars thus succeed for discrete invariant signatures, but fail when targets exhibit broad morphological diversity.

\section{Discussion}

\begin{table}[!htbp]
    \centering
    \scriptsize
    \setlength{\tabcolsep}{3.5pt}
    \caption{\textbf{Representative domain-specialist models vs.\ frontier VLMs.} Contextual reference baselines for each task modality. Specialist models are trained and evaluated on bespoke, task-specific pipelines; frontier VLM figures report zero-shot accuracy under the baseline guided prompt.}
    \label{tab:specialist}
    \begin{tabularx}{\textwidth}{lllXcc}
        \toprule
        Task & Modality & Specialist Method & Source Dataset & Specialist Metric & Best VLM (Zero-Shot) \\
        \midrule
        Task~1 & Optical images & Anomaly Det.\ / CNN~\cite{EuclidCollaboration2025b,Li2021} & Euclid / HSC & Precision: 59\% / 92\% & 74.5\% (Gemini~3~Pro) \\
        Task~2 & Radio maps & Equivariant CNN / SSL~\cite{Cecconello2025,Scaife2021} & MiraBest (FIRST) & Accuracy: 91.0\% & 84.6\% (Gemini~3~Pro) \\
        Task~3 & Multi-band SEDs & Tabular Random Forest~\cite{Zeraatgari2023} & AKARI / WISE & Precision: $>$98\% & 57.7\% (Gemini~3~Pro) \\
        Task~4 & Light curves & Avocado (LightGBM)~\cite{Boone2019} & PLAsTiCC & Accuracy: 94\%--100\% & 61.3\% (Gemini~3~Pro) \\
        Task~5 (Q2) & Optical spectra & Automated line fitting~\cite{Moustakas2023} & DESI DR1 & SNR $>5$ Rec.: $>$99\% & 87.4\% (Gemini~3~Pro) \\
        \bottomrule
    \end{tabularx}
\end{table}

\begin{figure}[!htbp]
    \centering
    \includegraphics[width=0.82\textwidth]{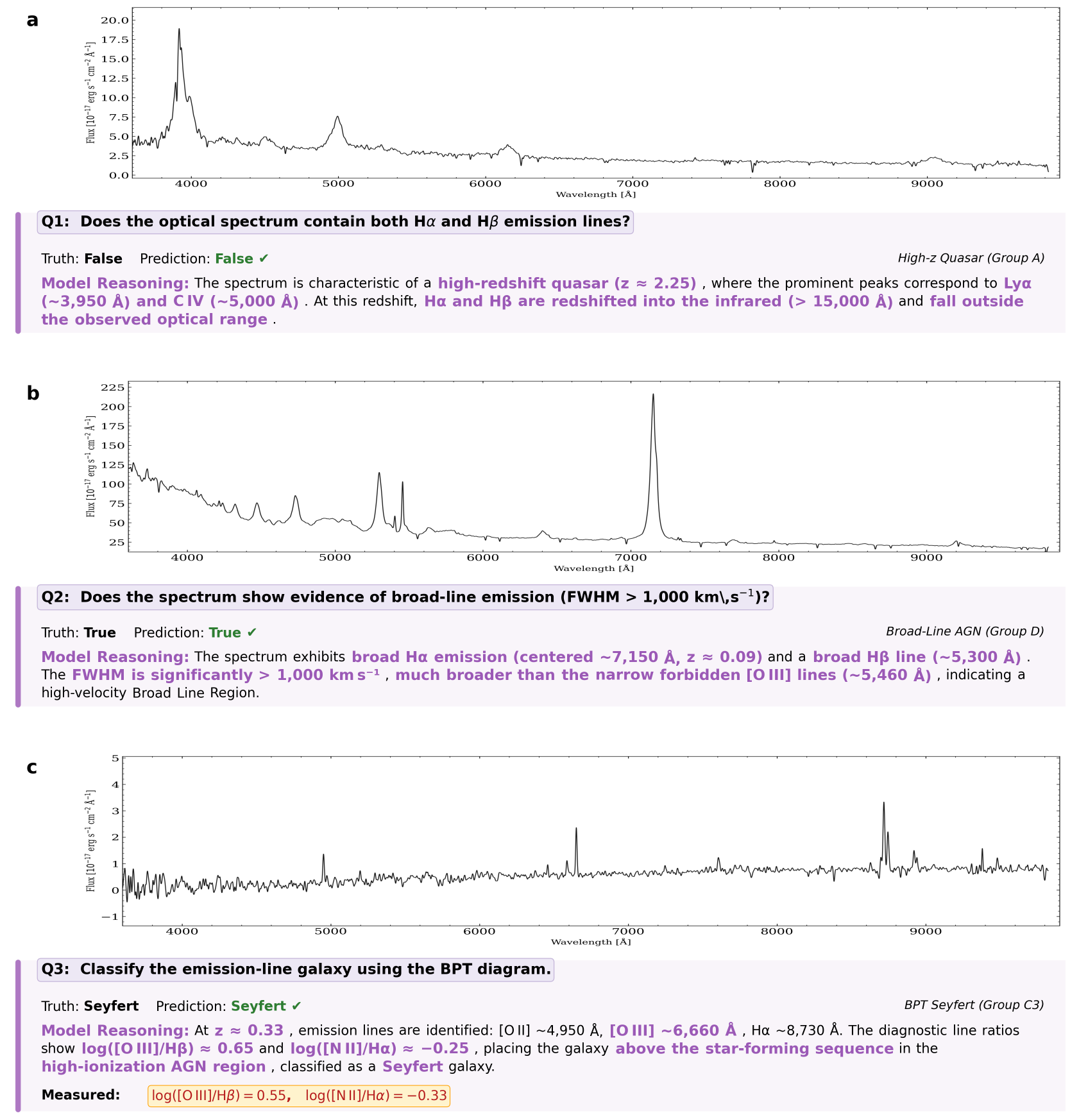}
    \caption{\textbf{Representative examples of physically valid reasoning by Gemini~3~Pro on Task~5 (Spectral Interpretation).} The three panels illustrate the hierarchical Q1$\to$Q2$\to$Q3 reasoning chain on DESI optical spectra, with the model's answer and verbatim reasoning shown below each input spectrum. Key reasoning steps are highlighted in bold purple.}
    \label{fig:reasoning}
\end{figure}

\noindent \textbf{Zero-shot VLMs versus domain-specialized astronomical models.}
Frontier VLMs achieve competitive zero-shot accuracy on spatial-morphology tasks but trail domain-specialized models (Table~\ref{tab:specialist}). In radio imaging (Task~2), Gemini~3~Pro (84.6\%) surpasses conventional ImageNet-transfer baselines (73.5\%) but trails self-supervised representations trained directly on radio data (91.0\%; \cite{Cecconello2025,Scaife2021}); on optical imaging (Task~1), top VLM precision is comparable to early anomaly detection ($\sim$18\%--59\%; \cite{EuclidCollaboration2025b}). The gap widens on tasks governed by precise numerical relationships: supervised learning on tabular photometry reaches $>98\%$ precision/recall on SED classification~\cite{Zeraatgari2023}, and feature-engineered models achieve 94\%--100\% accuracy on light-curve classes~\cite{Boone2019}, far exceeding zero-shot VLMs even with tabular inputs. Specialist models provide contextual reference rather than direct in-distribution comparisons: specialist algorithms operate on bespoke pipelines with supervised training, whereas frontier VLMs are evaluated zero-shot across unified representations without task tuning.

\noindent \textbf{Modality-dependent performance reveals task-specific strengths, not general competence.}
Gemini~3~Pro is the most consistent model, leading on Tasks~1 (74.5\%), 2 FIRST (84.6\%), 3 (57.7\%), 4 (61.3\%), and harder spectral subtasks (Q2 87.4\%, Q3 55.8\%). Yet task-specific strengths persist (Claude Opus 4.5 leads Q1 at 82.0\%), and overall performance is uncorrelated with general benchmark rankings. Paired randomization tests with Holm-Bonferroni correction confirm that while Gemini significantly leads on Tasks~1, 2 (FIRST), 4, and 5 (Q2, Q3), models are statistically indistinguishable on Task~3 and lower-resolution Task~2 NVSS. Astronomical capability is not a monolithic attribute: it reflects the interplay between visual encoder pretraining and the perceptual demands of each modality.

\noindent \textbf{Physical grounding and data representation jointly constrain astronomical inference.}
The disconnect between textual domain knowledge and applied reasoning is striking: while scoring $\sim$85\% on text-based astronomy Q\&A~\cite{Ting2024}, Claude Opus 4.5 and Qwen3 exhibit catastrophic mode collapses on real observational tasks (Tasks~2, 4, and 5 Q2/Q3). Our cross-model ablations show that non-imaging performance depends critically on data format: tabular light curves dramatically improve accuracy across all models (Gemini +13.3 pp, GPT-5.2 +13.4 pp, Claude +22.5 pp), while shuffled tables degrade performance, demonstrating that models actively exploit sequential numerical structure. Conversely, prompt grounding is model-dependent: physical prompts benefit Gemini and GPT-5.2 on imaging by balancing decision boundaries, whereas phenomenological prompts excel on light curves by explicitly articulating cadence artifacts.

\noindent \textbf{Autonomous physical reasoning versus fragile logic.}
Accuracy alone does not guarantee scientific reliability. To evaluate reasoning fidelity quantitatively, an astronomy expert evaluated a random sample of correctly classified Task~5 instances across models: an average of 26\% of correct predictions were accompanied by physically invalid or unprincipled reasoning chains (e.g., misidentifying emission lines, invoking incorrect atomic transitions, or asserting unphysical line intensity ratios). When reasoning succeeds, frontier VLMs exhibit impressive autonomy (Figure~\ref{fig:reasoning}): Gemini~3~Pro infers high-redshift quasars ($z \approx 2.25$) from unprompted Ly$\alpha$ and C~IV lines, contrasts broad Balmer profiles with narrow [O~III] lines to verify Broad Line Regions, and estimates line fluxes accurately on the BPT diagram. When models fail, they often latch onto local peaks with flawed justifications, creating a false impression of understanding that can mislead downstream pipelines (extended case excerpts and audit details in Appendix~\ref{sec:supp-reasoning}).

\noindent \textbf{Limitations.} (1)~\emph{Model scope:} We evaluate six frontier models, including two open-weight baselines (Qwen3-235B, Intern-S1-Pro); existing astronomy-fine-tuned models remain restricted to text or single modalities rather than cross-modal VLMs, leaving unified multi-modal fine-tuning as an important future direction. (2)~\emph{Sample size and rare objects:} Cohort sizes for Task~3 ($N=168$) and Task~4 ($N=142$) reflect the physical scarcity of complete panchromatic coverage ($\sim$5\% Type-2, $<1\%$ Type-1 AGN~\cite{Asmus2020}) and the scarcity of rare transients and template classes in simulated catalogs (e.g., confirmed optical TDEs $\sim$200 in literature~\cite{Langis2026}; well-sampled Miras $N=22$ in PLAsTiCC); balanced subsets prevent class imbalance artifacts, with bootstrap CIs and randomization tests making uncertainty explicit. (3)~\emph{Data contamination:} Public survey data (DESI, HSC, MiraBest, PLAsTiCC) may appear in web pretraining; inputs were strictly de-identified (removing coordinates, names, catalog IDs) to test observational interpretation rather than retrieval, though pretraining priors remain an inherent challenge for commercial foundation models. (4)~\emph{Reasoning evaluation:} Expert validation was conducted on Task~5; scalable automated verification across modalities remains an open problem.

\section{Conclusion}
\label{sec:conclusion}

\OURBENCH{} evaluates frontier VLMs on 4{,}100 expert-verified instances across five astronomical modalities, centered on AGN classification. Performance is strongly modality-dependent and uncorrelated with general benchmark rankings; all six models trail domain-specialized baselines on tasks governed by precise numerical relationships. Cross-model ablations uncover three bottlenecks: prompt grounding is model- and task-dependent, numerical tables outperform rendered plots by up to 23~pp on time-domain data, and 26\% of correct spectral classifications involve physically invalid reasoning. \OURBENCH{} provides the first systematic multi-modal VLM baseline in observational astronomy, clarifying the challenges for autonomous scientific deployment.

\clearpage

\begin{ack}
This work was supported by the National Natural Science Foundation of China (NSFC; Grant No.~12633003) and the National Key R\&D Program of China (Grant No.~2023YFA1607903).
\mbox{Hengxiao Guo} acknowledges additional support from the NSFC (Grant Nos.~12473018 and 12522304) and the Overseas Center Platform Projects of the Chinese Academy of Sciences (CAS; Grant No.~178GJHZ2023184MI).
\mbox{Wenwen Zuo} acknowledges support from the Shanghai Natural Science Foundation Youth Project (Grant No.~25ZR1402546), the Strategic Priority Research Program of CAS (Grant No.~XDB0800302), and the National Key R\&D Program of China (Grant No.~2025YFA1614102).
\end{ack}

\bibliographystyle{unsrtnat}
\bibliography{reference}

@String{iclr   = {Proceedings of the International Conference on Learning Representations}}

@Article{Aihara2017,
  author       = {Aihara, Hiroaki and Arimoto, Nobuo and Armstrong, Robert and Arnouts, Stéphane and Bahcall, Neta A and Bickerton, Steven and Bosch, James and Bundy, Kevin and Capak, Peter L and Chan, James H H and Chiba, Masashi and Coupon, Jean and Egami, Eiichi and Enoki, Motohiro and Finet, Francois and Fujimori, Hiroki and Fujimoto, Seiji and Furusawa, Hisanori and Furusawa, Junko and Goto, Tomotsugu and Goulding, Andy and Greco, Johnny P and Greene, Jenny E and Gunn, James E and Hamana, Takashi and Harikane, Yuichi and Hashimoto, Yasuhiro and Hattori, Takashi and Hayashi, Masao and Hayashi, Yusuke and Hełminiak, Krzysztof G and Higuchi, Ryo and Hikage, Chiaki and Ho, Paul T P and Hsieh, Bau-Ching and Huang, Kuiyun and Huang, Song and Ikeda, Hiroyuki and Imanishi, Masatoshi and Inoue, Akio K and Iwasawa, Kazushi and Iwata, Ikuru and Jaelani, Anton T and Jian, Hung-Yu and Kamata, Yukiko and Karoji, Hiroshi and Kashikawa, Nobunari and Katayama, Nobuhiko and Kawanomoto, Satoshi and Kayo, Issha and Koda, Jin and Koike, Michitaro and Kojima, Takashi and Komiyama, Yutaka and Konno, Akira and Koshida, Shintaro and Koyama, Yusei and Kusakabe, Haruka and Leauthaud, Alexie and Lee, Chien-Hsiu and Lin, Lihwai and Lin, Yen-Ting and Lupton, Robert H and Mandelbaum, Rachel and Matsuoka, Yoshiki and Medezinski, Elinor and Mineo, Sogo and Miyama, Shoken and Miyatake, Hironao and Miyazaki, Satoshi and Momose, Rieko and More, Anupreeta and More, Surhud and Moritani, Yuki and Moriya, Takashi J and Morokuma, Tomoki and Mukae, Shiro and Murata, Ryoma and Murayama, Hitoshi and Nagao, Tohru and Nakata, Fumiaki and Niida, Mana and Niikura, Hiroko and Nishizawa, Atsushi J and Obuchi, Yoshiyuki and Oguri, Masamune and Oishi, Yukie and Okabe, Nobuhiro and Okamoto, Sakurako and Okura, Yuki and Ono, Yoshiaki and Onodera, Masato and Onoue, Masafusa and Osato, Ken and Ouchi, Masami and Price, Paul A and Pyo, Tae-Soo and Sako, Masao and Sawicki, Marcin and Shibuya, Takatoshi and Shimasaku, Kazuhiro and Shimono, Atsushi and Shirasaki, Masato and Silverman, John D and Simet, Melanie and Speagle, Joshua and Spergel, David N and Strauss, Michael A and Sugahara, Yuma and Sugiyama, Naoshi and Suto, Yasushi and Suyu, Sherry H and Suzuki, Nao and Tait, Philip J and Takada, Masahiro and Takata, Tadafumi and Tamura, Naoyuki and Tanaka, Manobu M and Tanaka, Masaomi and Tanaka, Masayuki and Tanaka, Yoko and Terai, Tsuyoshi and Terashima, Yuichi and Toba, Yoshiki and Tominaga, Nozomu and Toshikawa, Jun and Turner, Edwin L and Uchida, Tomohisa and Uchiyama, Hisakazu and Umetsu, Keiichi and Uraguchi, Fumihiro and Urata, Yuji and Usuda, Tomonori and Utsumi, Yousuke and Wang, Shiang-Yu and Wang, Wei-Hao and Wong, Kenneth C and Yabe, Kiyoto and Yamada, Yoshihiko and Yamanoi, Hitomi and Yasuda, Naoki and Yeh, Sherry and Yonehara, Atsunori and Yuma, Suraphong},
  journal      = {Publications of the Astronomical Society of Japan},
  title        = {The Hyper Suprime-Cam SSP Survey: Overview and survey design},
  year         = {2017},
  issn         = {2053-051X},
  month        = sep,
  number       = {SP1},
  volume       = {70},
  creationdate = {2026-02-23T02:49:07},
  doi          = {10.1093/pasj/psx066},
  publisher    = {Oxford University Press (OUP)},
}

@Article{Aihara2022,
  author       = {Aihara, Hiroaki and AlSayyad, Yusra and Ando, Makoto and Armstrong, Robert and Bosch, James and Egami, Eiichi and Furusawa, Hisanori and Furusawa, Junko and Harasawa, Sumiko and Harikane, Yuichi and Hsieh, Bau-Ching and Ikeda, Hiroyuki and Ito, Kei and Iwata, Ikuru and Kodama, Tadayuki and Koike, Michitaro and Kokubo, Mitsuru and Komiyama, Yutaka and Li, Xiangchong and Liang, Yongming and Lin, Yen-Ting and Lupton, Robert H and Lust, Nate B and MacArthur, Lauren A and Mawatari, Ken and Mineo, Sogo and Miyatake, Hironao and Miyazaki, Satoshi and More, Surhud and Morishima, Takahiro and Murayama, Hitoshi and Nakajima, Kimihiko and Nakata, Fumiaki and Nishizawa, Atsushi J and Oguri, Masamune and Okabe, Nobuhiro and Okura, Yuki and Ono, Yoshiaki and Osato, Ken and Ouchi, Masami and Pan, Yen-Chen and Plazas Malagón, Andrés A and Price, Paul A and Reed, Sophie L and Rykoff, Eli S and Shibuya, Takatoshi and Simunovic, Mirko and Strauss, Michael A and Sugimori, Kanako and Suto, Yasushi and Suzuki, Nao and Takada, Masahiro and Takagi, Yuhei and Takata, Tadafumi and Takita, Satoshi and Tanaka, Masayuki and Tang, Shenli and Taranu, Dan S and Terai, Tsuyoshi and Toba, Yoshiki and Turner, Edwin L and Uchiyama, Hisakazu and Vijarnwannaluk, Bovornpratch and Waters, Christopher Z and Yamada, Yoshihiko and Yamamoto, Naoaki and Yamashita, Takuji},
  journal      = {Publications of the Astronomical Society of Japan},
  title        = {Third data release of the Hyper Suprime-Cam Subaru Strategic Program},
  year         = {2022},
  issn         = {2053-051X},
  month        = feb,
  number       = {2},
  pages        = {247--272},
  volume       = {74},
  creationdate = {2026-02-23T02:50:13},
  doi          = {10.1093/pasj/psab122},
  publisher    = {Oxford University Press (OUP)},
}

@Article{Li2021,
  author       = {Li, Junyao and Silverman, John D. and Ding, Xuheng and Strauss, Michael A. and Goulding, Andy and Birrer, Simon and Yesuf, Hassen M. and Xue, Yongquan and Kawinwanichakij, Lalitwadee and Matsuoka, Yoshiki and Toba, Yoshiki and Nagao, Tohru and Schramm, Malte and Inayoshi, Kohei},
  journal      = {The Astrophysical Journal},
  title        = {The Sizes of Quasar Host Galaxies in the Hyper Suprime-Cam Subaru Strategic Program},
  year         = {2021},
  issn         = {1538-4357},
  month        = aug,
  number       = {1},
  pages        = {22},
  volume       = {918},
  creationdate = {2026-02-23T02:51:22},
  doi          = {10.3847/1538-4357/ac06a8},
  publisher    = {American Astronomical Society},
}

@Article{Simard2011,
  author       = {Simard, Luc and Trevor Mendel, J. and Patton, David R. and Ellison, Sara L. and McConnachie, Alan W.},
  journal      = {The Astrophysical Journal Supplement Series},
  title        = {A CATALOG OF BULGE+DISK DECOMPOSITIONS AND UPDATED PHOTOMETRY FOR 1.12 MILLION GALAXIES IN THE SLOAN DIGITAL SKY SURVEY},
  year         = {2011},
  issn         = {1538-4365},
  month        = aug,
  number       = {1},
  pages        = {11},
  volume       = {196},
  creationdate = {2026-02-23T02:52:11},
  doi          = {10.1088/0067-0049/196/1/11},
  publisher    = {American Astronomical Society},
}

@Misc{Coe2012,
  author        = {{Coe}, D.},
  month         = aug,
  title         = {{Trilogy: Image composition software}},
  year          = {2012},
  adsurl        = {https://ui.adsabs.harvard.edu/abs/2012ascl.soft08009C},
  archiveprefix = {ascl},
  creationdate  = {2026-02-23T02:53:01},
  eid           = {ascl:1508.009},
  eprint        = {1508.009},
  pages         = {ascl:1508.009},
}

@ARTICLE{SKA2013,
       author = {{Aharonian}, F. and {Arshakian}, T.~G. and {Allen}, B. and {Banerjee}, R. and {Beck}, R. and {Becker}, W. and {Bomans}, D.~J. and {Breitschwerdt}, D. and {Br{\"u}ggen}, M. and {Brunthaler}, A. and {Catinella}, B. and {Champion}, D. and {Ciardi}, B. and {Crocker}, R. and {de Avillez}, M.~A. and {Dettmar}, R.~J. and {Engels}, D. and {En{\ss}lin}, T. and {Enke}, H. and {Fieseler}, T. and {Gizon}, L. and {Hackmann}, E. and {Hartmann}, B. and {Henkel}, C. and {Hoeft}, M. and {Iapichino}, L. and {Innes}, D. and {James}, C. and {Jasche}, J. and {Jones}, D. and {Kagramanova}, V. and {Kauffmann}, G. and {Keane}, E. and {Kerp}, J. and {Kl{\"o}ckner}, H.-R. and {Kokkotas}, K. and {Kramer}, M. and {Krause}, M. and {Krause}, M. and {Krupp}, N. and {Kunz}, J. and {L{\"a}mmerzahl}, C. and {Lee}, K.~J. and {List}, M. and {Liu}, K. and {Lobanov}, A. and {Mann}, G. and {Merloni}, A. and {Middelberg}, E. and {Niemeyer}, J. and {Noutsos}, A. and {Perlick}, V. and {Reich}, W. and {Richter}, P. and {Roy}, A. and {Saintonge}, A. and {Sch{\"a}fer}, G. and {Schaffner-Bielich}, J. and {Schinnerer}, E. and {Schleicher}, D. and {Schneider}, P. and {Schwarz}, D.~J. and {Sedrakian}, A. and {Sesana}, A. and {Smol{\v{c}}i{\'c}}, V. and {Solanki}, S. and {Tuffs}, R. and {Vetter}, M. and {Weber}, E. and {Weller}, J. and {Wex}, N. and {Wucknitz}, O. and {Zwaan}, M.},
        title = "{Pathway to the Square Kilometre Array - The German White Paper -}",
      journal = {arXiv e-prints},
         year = 2013,
        month = jan,
          eid = {arXiv:1301.4124},
        pages = {arXiv:1301.4124},
          doi = {10.48550/arXiv.1301.4124},
archivePrefix = {arXiv},
       eprint = {1301.4124},
 primaryClass = {astro-ph.IM},
       adsurl = {https://ui.adsabs.harvard.edu/abs/2013arXiv1301.4124A}
}

@Article{Parker2025,
  author        = {Parker, Liam and Lanusse, Francois and Shen, Jeff and Liu, Ollie and Hehir, Tom and Sarra, Leopoldo and Meyer, Lucas and Bowles, Micah and Wagner-Carena, Sebastian and Qu, Helen and Golkar, Siavash and Bietti, Alberto and Bourfoune, Hatim and Casserau, Nathan and Cornette, Pierre and Hirashima, Keiya and Krawezik, Geraud and Ohana, Ruben and Lourie, Nicholas and McCabe, Michael and Morel, Rudy and Mukhopadhyay, Payel and Pettee, Mariel and Blancard, Bruno Regaldo-Saint and Cho, Kyunghyun and Cranmer, Miles and Ho, Shirley},
  title         = {AION-1: Omnimodal Foundation Model for Astronomical Sciences},
  year          = {2025},
  month         = oct,
  archiveprefix = {arXiv},
  copyright     = {Creative Commons Attribution 4.0 International},
  creationdate  = {2026-01-26T13:14:11},
  doi           = {10.48550/ARXIV.2510.17960},
  eprint        = {2510.17960},
  primaryclass  = {astro-ph.IM},
  publisher     = {arXiv},
}

@Article{Jiang2026,
  author        = {Jiang, Yuanzhe and Shen, Yue and Merz, Grant and Lin, Shurui and Liu, Xin and Pan, Zhiwei and Zhuang, Mingyang and Roster, William and Salvato, Mara and Siudek, Malgorzata and Stevens, Grant},
  title         = {DeepDISC-Euclid: Source Classification and Photometric Redshifts in Euclid Deep Field North With a Pixel-Level Deep Learning Approach},
  year          = {2026},
  month         = apr,
  archiveprefix = {arXiv},
  copyright     = {Creative Commons Attribution 4.0 International},
  creationdate  = {2026-04-06T11:19:09},
  doi           = {10.48550/ARXIV.2604.03182},
  eprint        = {2604.03182},
  primaryclass  = {astro-ph.GA},
  publisher     = {arXiv},
}

@ARTICLE{Roman2025,
       author = {{Observations Time Allocation Committee}, Roman and {Community Survey Definition Committees}, Core},
        title = "{Roman Observations Time Allocation Committee: Final Report and Recommendations}",
      journal = {arXiv e-prints},
         year = 2025,
        month = may,
          eid = {arXiv:2505.10574},
        pages = {arXiv:2505.10574},
          doi = {10.48550/arXiv.2505.10574},
archivePrefix = {arXiv},
       eprint = {2505.10574},
 primaryClass = {astro-ph.IM},
       adsurl = {https://ui.adsabs.harvard.edu/abs/2025arXiv250510574O}
}

@Article{Porter2023,
  author       = {Porter, Fiona A M and Scaife, Anna M M},
  journal      = {RAS Techniques and Instruments},
  title        = {MiraBest: a data set of morphologically classified radio galaxies for machine learning},
  year         = {2023},
  issn         = {2752-8200},
  month        = jan,
  number       = {1},
  pages        = {293--306},
  volume       = {2},
  creationdate = {2026-02-23T04:02:09},
  doi          = {10.1093/rasti/rzad017},
  publisher    = {Oxford University Press (OUP)},
}

@Article{Cecconello2025,
  author        = {Cecconello, Thomas and Riggi, Simone and Becciani, Ugo and Vitello, Fabio and Hopkins, Andrew M. and Vizzari, Giuseppe and Spampinato, Concetto and Palazzo, Simone},
  title         = {Self-Supervised Learning for Radio-Astronomy Source Classification: A Benchmark},
  year          = {2025},
  issn          = {1611-3349},
  month         = nov,
  pages         = {424--439},
  archiveprefix = {arXiv},
  booktitle     = {Pattern Recognition. ICPR 2024 International Workshops and Challenges},
  copyright     = {arXiv.org perpetual, non-exclusive license},
  creationdate  = {2026-04-08T17:30:36},
  doi           = {10.1007/978-3-031-88217-3_32},
  eprint        = {2411.14078},
  isbn          = {9783031882173},
  primaryclass  = {astro-ph.IM},
  publisher     = {Springer Nature Switzerland},
}

@Article{Becker1995,
  author       = {Becker, Robert H. and White, Richard L. and Helfand, David J.},
  journal      = {The Astrophysical Journal},
  title        = {The FIRST Survey: Faint Images of the Radio Sky at Twenty Centimeters},
  year         = {1995},
  issn         = {1538-4357},
  month        = sep,
  pages        = {559},
  volume       = {450},
  creationdate = {2026-02-23T04:05:25},
  doi          = {10.1086/176166},
  publisher    = {American Astronomical Society},
}

@Article{Condon1998,
  author       = {Condon, J. J. and Cotton, W. D. and Greisen, E. W. and Yin, Q. F. and Perley, R. A. and Taylor, G. B. and Broderick, J. J.},
  journal      = {The Astronomical Journal},
  title        = {The NRAO VLA Sky Survey},
  year         = {1998},
  issn         = {0004-6256},
  month        = may,
  number       = {5},
  pages        = {1693--1716},
  volume       = {115},
  creationdate = {2026-02-23T04:05:38},
  doi          = {10.1086/300337},
  publisher    = {American Astronomical Society},
}

@Article{Kim2020,
  author       = {Kim, Seong Jin and Oi, Nagisa and Goto, Tomotsugu and Ikeda, Hiroyuki and Ho, Simon C-C and Shim, Hyunjin and Toba, Yoshiki and Hwang, Ho Seong and Hashimoto, Tetsuya and Barrufet, Laia and Malkan, Matthew and Kim, Helen K and Huang, Ting-Chi and Matsuhara, Hideo and Miyaji, Takamitsu and Pearson, Chris and Serjeant, Stephen and Santos, Daryl Joe D and Kim, Eunbin and Pollo, Agnieszka and Jeong, Woong-Seob and Wang, Ting-Wen and Momose, Rieko and Takagi, Toshinobu},
  journal      = {Monthly Notices of the Royal Astronomical Society},
  title        = {Identification of AKARI infrared sources by the Deep HSC Optical Survey: construction of a new band-merged catalogue in the North Ecliptic Pole Wide field},
  year         = {2020},
  issn         = {1365-2966},
  month        = dec,
  number       = {3},
  pages        = {4078--4094},
  volume       = {500},
  creationdate = {2026-02-23T23:50:00},
  doi          = {10.1093/mnras/staa3359},
  publisher    = {Oxford University Press (OUP)},
}

@Software{Moustakas2023,
  adsurl        = {https://ui.adsabs.harvard.edu/abs/2023ascl.soft08005M},
  archiveprefix = {ascl},
  author        = {{Moustakas}, John and {Buhler}, Jeremy and {Scholte}, Dirk and {Dey}, Biprateep and {Khederlarian}, Ashod},
  creationdate  = {2026-02-24T02:42:35},
  eid           = {ascl:2308.005},
  eprint        = {2308.005},
  howpublished  = {Astrophysics Source Code Library, record ascl:2308.005},
  month         = aug,
  title         = {{FastSpecFit: Fast spectral synthesis and emission-line fitting of DESI spectra}},
  year          = {2023},
}

@Article{Baldwin1981,
  author       = {Baldwin, J. A. and Phillips, M. M. and Terlevich, R.},
  journal      = {Publications of the Astronomical Society of the Pacific},
  title        = {Classification parameters for the emission-line spectra of extragalactic objects},
  year         = {1981},
  issn         = {1538-3873},
  month        = feb,
  pages        = {5},
  volume       = {93},
  creationdate = {2026-02-24T02:44:03},
  doi          = {10.1086/130766},
  publisher    = {IOP Publishing},
}

@Article{SanchezSaez2021,
  author       = {Sánchez-Sáez, P. and Reyes, I. and Valenzuela, C. and Förster, F. and Eyheramendy, S. and Elorrieta, F. and Bauer, F. E. and Cabrera-Vives, G. and Estévez, P. A. and Catelan, M. and Pignata, G. and Huijse, P. and De Cicco, D. and Arévalo, P. and Carrasco-Davis, R. and Abril, J. and Kurtev, R. and Borissova, J. and Arredondo, J. and Castillo-Navarrete, E. and Rodriguez, D. and Ruz-Mieres, D. and Moya, A. and Sabatini-Gacitúa, L. and Sepúlveda-Cobo, C. and Camacho-Iñiguez, E.},
  journal      = {The Astronomical Journal},
  title        = {Alert Classification for the ALeRCE Broker System: The Light Curve Classifier},
  year         = {2021},
  issn         = {1538-3881},
  month        = feb,
  number       = {3},
  pages        = {141},
  volume       = {161},
  creationdate = {2026-02-24T03:31:59},
  doi          = {10.3847/1538-3881/abd5c1},
  publisher    = {American Astronomical Society},
}

@Article{DESICollaboration2025,
  author        = {{DESI Collaboration} and Abdul-Karim, M. and Adame, A. G. and Aguado, D. and Aguilar, J. and Ahlen, S. and Alam, S. and Aldering, G. and Alexander, D. M. and Alfarsy, R. and Allen, L. and Prieto, C. Allende and Alves, O. and Anand, A. and Andrade, U. and Armengaud, E. and Avila, S. and Aviles, A. and Awan, H. and Bailey, S. and Lizancos, A. Baleato and Ballester, O. and Bault, A. and Bautista, J. and BenZvi, S. and Silva, L. Beraldo e and Bermejo-Climent, J. R. and Beutler, F. and Bianchi, D. and Blake, C. and Blum, R. and Bolton, A. S. and Bonici, M. and Brieden, S. and Brodzeller, A. and Brooks, D. and Buckley-Geer, E. and Burtin, E. and Canning, R. and Rosell, A. Carnero and Carr, A. and Carrilho, P. and Casas, L. and Castander, F. J. and Cereskaite, R. and Cervantes-Cota, J. L. and Chaussidon, E. and Chaves-Montero, J. and Chen, S. and Chen, X. and Claybaugh, T. and Cole, S. and Cooper, A. P. and Cousinou, M. -C. and Cuceu, A. and Davis, T. M. and Dawson, K. S. and de Belsunce, R. and de la Cruz, R. and de la Macorra, A. and de Mattia, A. and Deiosso, N. and Della Costa, J. and Demina, R. and Demirbozan, U. and DeRose, J. and Dey, A. and Dey, B. and Ding, J. and Ding, Z. and Doel, P. and Douglass, K. and Dowicz, M. and Ebina, H. and Edelstein, J. and Eisenstein, D. J. and Elbers, W. and Emas, N. and Escoffier, S. and Fagrelius, P. and Fan, X. and Fanning, K. and Fawcett, V. A. and Fernández-García, E. and Ferraro, S. and Findlay, N. and Font-Ribera, A. and Forero-Romero, J. E. and Forero-Sánchez, D. and Frenk, C. S. and Gänsicke, B. T. and Galbany, L. and García-Bellido, J. and Garcia-Quintero, C. and Garrison, L. H. and Gaztañaga, E. and Gil-Marín, H. and Gnedin, O. Y. and Gontcho, S. Gontcho A and Gonzalez-Morales, A. X. and Gonzalez-Perez, V. and Gordon, C. and Graur, O. and Green, D. and Gruen, D. and Gsponer, R. and Guandalin, C. and Gutierrez, G. and Guy, J. and Hahn, C. and Han, J. J. and Han, J. and He, S. and Herrera-Alcantar, H. K. and Honscheid, K. and Hou, J. and Howlett, C. and Huterer, D. and Iršič, V. and Ishak, M. and Jacques, A. and Jimenez, J. and Jing, Y. P. and Joachimi, B. and Joudaki, S. and Joyce, R. and Jullo, E. and Juneau, S. and Karaçaylı, N. G. and Karim, T. and Kehoe, R. and Kent, S. and Khederlarian, A. and Kirkby, D. and Kisner, T. and Kitaura, F. -S. and Kizhuprakkat, N. and Kong, H. and Koposov, S. E. and Kremin, A. and Krolewski, A. and Lahav, O. and Lai, Y. and Lamman, C. and Lan, T. -W. and Landriau, M. and Lang, D. and Lange, J. U. and Lasker, J. and Goff, J. M. Le and Guillou, L. Le and Leauthaud, A. and Levi, M. E. and Li, S. and Li, T. S. and Lodha, K. and Lokken, M. and Luo, Y. and Magneville, C. and Manera, M. and Manser, C. J. and Margala, D. and Martini, P. and Maus, M. and McCullough, J. and McDonald, P. and Medina, G. E. and Medina-Varela, L. and Meisner, A. and Mena-Fernández, J. and Menegas, A. and Mezcua, M. and Miquel, R. and Montero-Camacho, P. and Moon, J. and Moustakas, J. and Muñoz-Gutiérrez, A. and Muñoz-Santos, D. and Myers, A. D. and Myles, J. and Nadathur, S. and Najita, J. and Napolitano, L. and Newman, J. A. and Nikakhtar, F. and Nikutta, R. and Niz, G. and Noriega, H. E. and Padmanabhan, N. and Paillas, E. and Palanque-Delabrouille, N. and Palmese, A. and Pan, J. and Pan, Z. and Parkinson, D. and Peacock, J. and Percival, W. J. and Pérez-Fernández, A. and Pérez-Ràfols, I. and Peterson, P. and Piat, J. and Pieri, M. M. and Pinon, M. and Poppett, C. and Porredon, A. and Prada, F. and Pucha, R. and Qin, F. and Rabinowitz, D. and Raichoor, A. and Ramírez-Pérez, C. and Ramirez-Solano, S. and Rashkovetskyi, M. and Ravoux, C. and Riley, A. H. and Rocher, A. and Rockosi, C. and Rohlf, J. and Ross, A. J. and Rossi, G. and Ruggeri, R. and Ruhlmann-Kleider, V. and Sabiu, C. G. and Said, K. and Saintonge, A. and Samushia, L. and Sanchez, E. and Sanders, N. and Saulder, C. and Schlafly, E. F. and Schlegel, D. and Scholte, D. and Schubnell, M. and Seo, H. and Shafieloo, A. and Sharples, R. and Silber, J. and Siudek, M. and Smith, A. and Sprayberry, D. and Suárez-Pérez, J. and Swanson, J. and Tan, T. and Tarlé, G. and Taylor, P. and Thomas, G. and Tojeiro, R. and Turner, R. J. and Turner, W. and Ureña-López, L. A. and Vaisakh, R. and Valluri, M. and Vargas-Magaña, M. and Verde, L. and Walther, M. and Wang, B. and Wang, M. S. and Wang, W. and Weaver, B. A. and Weaverdyck, N. and Wechsler, R. H. and White, M. and Wolfson, M. and Yang, J. and Yèche, C. and Youles, S. and Yu, J. and Yuan, S. and Zaborowski, E. A. and Zarrouk, P. and Zhang, H. and Zhao, C. and Zhao, R. and Zheng, Z. and Zhou, R. and Zou, H. and Zou, S. and Zu, Y.},
  title         = {Data Release 1 of the Dark Energy Spectroscopic Instrument},
  year          = {2025},
  month         = mar,
  archiveprefix = {arXiv},
  copyright     = {Creative Commons Attribution 4.0 International},
  creationdate  = {2026-02-24T03:36:10},
  doi           = {10.48550/ARXIV.2503.14745},
  eprint        = {2503.14745},
  primaryclass  = {astro-ph.CO},
  publisher     = {arXiv},
}

@Article{Rein2024,
  author        = {Rein, David and Hou, Betty Li and Stickland, Asa Cooper and Petty, Jackson and Pang, Richard Yuanzhe and Dirani, Julien and Michael, Julian and Bowman, Samuel R.},
  title         = {{GPQA}: A Graduate-Level Google-Proof Q\&A Benchmark},
  year          = {2024},
  eprint        = {2311.12022},
  archiveprefix = {arXiv},
  primaryclass  = {cs.CL},
  note          = {ICLR 2024 Spotlight},
}

@Article{RomeraParedes2024,
  author  = {Romera-Paredes, Bernardino and Barekatain, Mohammadamin and Novikov, Alexander and Balog, Matej and Kumar, M. Pawan and Dupont, Emilien and Ruiz, Francisco J. R. and Brockschmidt, Marc and Kohli, Pushmeet and others},
  title   = {Mathematical discoveries from program search with large language models},
  journal = {Nature},
  year    = {2024},
  volume  = {625},
  number  = {7995},
  pages   = {468--475},
  doi     = {10.1038/s41586-023-06924-6},
}

@Article{Phan2025,
  author        = {Phan, Long and Gatti, Alice and Han, Ziwen and Li, Nathaniel and others},
  title         = {Humanity's Last Exam},
  year          = {2025},
  eprint        = {2501.14249},
  archiveprefix = {arXiv},
  primaryclass  = {cs.CL},
}

@Article{Dieleman2015,
  author  = {Dieleman, Sander and Willett, Kyle W. and Dambre, Joni},
  title   = {Rotation-invariant convolutional neural networks for galaxy morphology prediction},
  journal = {Monthly Notices of the Royal Astronomical Society},
  year    = {2015},
  volume  = {450},
  number  = {2},
  pages   = {1441--1459},
  doi     = {10.1093/mnras/stv632},
}

@Article{Walmsley2022,
  author  = {Walmsley, Mike and Lintott, Chris and G{\'e}ron, Tobias and Kruk, Sandor and Krawczyk, Coleman and Willett, Kyle W. and Bamford, Steven and Kelvin, Lee S. and Fortson, Lucy and others},
  title   = {Galaxy Zoo DECaLS: Detailed visual morphology measurements from volunteers and deep learning for 314\,000 galaxies},
  journal = {Monthly Notices of the Royal Astronomical Society},
  year    = {2022},
  volume  = {509},
  number  = {3},
  pages   = {3966--3988},
  doi     = {10.1093/mnras/stab2093},
}

@dataset{plasticc2019,
  author       = {PLASTICC Team and
                  PLASTICC Modelers},
  title        = {Unblinded Data for PLAsTiCC Classification
                   Challenge
                  },
  month        = jan,
  year         = 2019,
  publisher    = {Zenodo},
  version      = {v1},
  doi          = {10.5281/zenodo.2539456},
  url          = {https://doi.org/10.5281/zenodo.2539456},
}

@Article{Zeraatgari2023,
  author        = {Zeraatgari, Fatemeh Zahra and Hafezianzadeh, Fatemeh and Zhang, Yanxia and Mei, Liquan and Ayubinia, Ashraf and Mosallanezhad, Amin and Zhang, Jingyi},
  journal       = {MNRAS},
  title         = {Machine learning-based photometric classification of galaxies, quasars, emission-line galaxies, and stars},
  year          = {2023},
  issn          = {1365-2966},
  month         = nov,
  number        = {3},
  pages         = {4677--4689},
  volume        = {527},
  archiveprefix = {arXiv},
  copyright     = {Creative Commons Attribution Non Commercial No Derivatives 4.0 International},
  creationdate  = {2026-04-09T00:11:02},
  doi           = {10.1093/mnras/stad3436},
  eprint        = {2311.02951},
  fjournal      = {Monthly Notices of the Royal Astronomical Society},
  primaryclass  = {astro-ph.GA},
  publisher     = {Oxford University Press (OUP)},
}

@Article{Moller2020,
  author  = {M{\"o}ller, Anais and de Boissi{\`e}re, Thibault},
  title   = {{SuperNNova}: an open-source framework for {Bayesian}, neural network-based supernova classification},
  journal = {Monthly Notices of the Royal Astronomical Society},
  year    = {2020},
  volume  = {491},
  number  = {3},
  pages   = {4277--4293},
  doi     = {10.1093/mnras/stz3312},
}

@Article{Wright2010,
  title={The Wide-field Infrared Survey Explorer (WISE): mission description and initial on-orbit performance},
  author={Wright, Edward L and Eisenhardt, Peter RM and Mainzer, Amy K and Ressler, Michael E and Cutri, Roc M and Jarrett, Thomas and Kirkpatrick, J Davy and Padgett, Deborah and McMillan, Robert S and Skrutskie, Michael and others},
  journal={The Astronomical Journal},
  volume={140},
  number={6},
  pages={1868--1881},
  year={2010},
  publisher={The American Astronomical Society}
}

@Article{Veilleux1987,
  author       = {Veilleux, Sylvain and Osterbrock, Donald E.},
  journal      = {ApJS},
  title        = {Spectral classification of emission-line galaxies},
  year         = {1987},
  issn         = {1538-4365},
  month        = feb,
  pages        = {295},
  volume       = {63},
  creationdate = {2024-09-23T12:02:01},
  doi          = {10.1086/191166},
  fjournal     = {The Astrophysical Journal Supplement Series},
  publisher    = {American Astronomical Society},
}

@Article{Boone2019,
  author       = {Boone, Kyle},
  journal      = {AJ},
  title        = {Avocado: Photometric Classification of Astronomical Transients with Gaussian Process Augmentation},
  year         = {2019},
  issn         = {1538-3881},
  month        = dec,
  number       = {6},
  pages        = {257},
  volume       = {158},
  creationdate = {2026-04-09T00:46:38},
  doi          = {10.3847/1538-3881/ab5182},
  fjournal     = {The Astronomical Journal},
  publisher    = {American Astronomical Society},
}

@Article{Boquien2019,
  author  = {Boquien, M{\'e}d{\'e}ric and Burgarella, Denis and Roehlly, Yannick and Buat, V{\'e}ronique and Ciesla, Laure and Corre, Denis and Iber, Alberto K. and Hatziminaoglou, Evanthia},
  title   = {{CIGALE}: a python Code Investigating {GALaxy} Emission},
  journal = {Astronomy \& Astrophysics},
  year    = {2019},
  volume  = {622},
  pages   = {A103},
  doi     = {10.1051/0004-6361/201834156},
}

@Article{Johnson2021,
  author  = {Johnson, Benjamin D. and Leja, Joel and Conroy, Charlie and Speagle, Joshua S.},
  title   = {Stellar Population Inference with {Prospector}},
  journal = {The Astrophysical Journal Supplement Series},
  year    = {2021},
  volume  = {254},
  number  = {2},
  pages   = {22},
  doi     = {10.3847/1538-4365/abef67},
}

@Article{Ivezic2019,
  author  = {Ivezi{\'c}, {\v{Z}}eljko and Kahn, Steven M. and Tyson, J. Anthony and Abel, Bob and Acosta, Emily and Allsman, Robyn and Alonso, David and AlSayyad, Yusra and Anderson, Scott F. and Andrew, John and others},
  title   = {{LSST}: From Science Drivers to Reference Design and Anticipated Data Products},
  journal = {The Astrophysical Journal},
  year    = {2019},
  volume  = {873},
  pages   = {111},
  doi     = {10.3847/1538-4357/ab042c},
}

@Article{Stoppa2025,
  author  = {Stoppa, F. and Bulmus, T. and others},
  title   = {Textual interpretation of transient image classifications from large language models},
  journal = {Nature Astronomy},
  year    = {2025},
  volume  = {9},
  pages   = {1869--1878},
  doi     = {10.1038/s41550-025-02670-z},
}

@Article{Sun2024,
  author        = {Sun, Zechang and Ting, Yuan-Sen and Liang, Yaobo and Duan, Nan and Huang, Song and Cai, Zheng},
  title         = {Interpreting Multi-band Galaxy Observations with Large Language Model-Based Agents},
  year          = {2024},
  eprint        = {2409.14807},
  archiveprefix = {arXiv},
  primaryclass  = {astro-ph.GA},
}

@Article{Riggi2025,
  author  = {Riggi, S. and Cecconello, T. and Pilzer, A. and others},
  title   = {{radio-llava}: Advancing Vision-Language Models for Radio Astronomical Source Analysis},
  journal = {Publications of the Astronomical Society of Australia},
  year    = {2025},
  doi     = {10.1017/pasa.2025.10082},
  eprint  = {2503.23859},
  archiveprefix = {arXiv},
}

@Article{Drozdova2025,
  author        = {Drozdova, Mariia and Lastufka, Erica and Kinakh, Vitaliy and others},
  title         = {Radio Astronomy in the Era of Vision-Language Models: Prompt Sensitivity and Adaptation},
  year          = {2025},
  eprint        = {2509.02615},
  archiveprefix = {arXiv},
  primaryclass  = {astro-ph.IM},
}

@Article{Ting2024,
  author        = {Ting, Yuan-Sen and Nguyen, Tuan Dung and Ghosal, Tirthankar and others},
  title         = {{AstroMLab} 1: Who Wins Astronomy Jeopardy!?},
  year          = {2024},
  eprint        = {2407.11194},
  archiveprefix = {arXiv},
  primaryclass  = {astro-ph.IM},
}

@Article{deHaan2025,
  author        = {de Haan, Tijmen and Ting, Yuan-Sen and Ghosal, Tirthankar and others},
  title         = {{AstroMLab} 3: Achieving {GPT-4o} Level Performance in Astronomy with a Specialized 8B-Parameter Large Language Model},
  journal       = {Scientific Reports},
  year          = {2025},
  volume        = {15},
  number        = {1},
  pages         = {13751},
  doi           = {10.1038/s41598-025-97131-y},
  eprint        = {2411.09012},
  archiveprefix = {arXiv},
}

@Article{Shi2025,
  author        = {Shi, Jie and Tang, Xin and Huang, Yi and Li, Yong and Kong, Xiao and Zhang, Yanxia and Yue, Chengze},
  title         = {{AstroMMBench}: A Benchmark for Evaluating Multimodal Large Language Models Capabilities in Astronomy},
  year          = {2025},
  eprint        = {2510.00063},
  archiveprefix = {arXiv},
  primaryclass  = {astro-ph.IM},
}

@Article{Bellm2019,
  author  = {Bellm, Eric C. and Kulkarni, Shrinivas R. and Graham, Matthew J. and Dekany, Richard and Smith, Roger M. and Riddle, Reed and others},
  title   = {The {Zwicky Transient Facility}: System Overview, Performance, and First Results},
  journal = {Publications of the Astronomical Society of the Pacific},
  year    = {2019},
  volume  = {131},
  pages   = {018002},
  doi     = {10.1088/1538-3873/aaecbe},
}

@article{Gottweis2025,
  author  = {Gottweis, Juraj and Cai, Wei and Huang, Adrian and Taly, Avinatan and Brenner, Michael P.},
  title   = {Towards an {AI} co-scientist},
  journal = {arXiv preprint arXiv:2502.18864},
  year    = {2025},
}

@article{GPT52,
  title   = {{OpenAI GPT-5 System Card}},
  author  = {{OpenAI}},
  journal = {arXiv preprint arXiv:2601.03267},
  year    = {2025},
  url     = {https://openai.com/index/introducing-gpt-5-2/},
}

@misc{ClaudeOpus45,
  title        = {{Claude Opus 4.5}},
  author       = {{Anthropic}},
  howpublished = {Anthropic Blog},
  year         = {2025},
  url          = {https://www.anthropic.com/news/claude-opus-4-5},
}

@misc{Gemini3Pro,
  title        = {{Gemini 3 Pro}},
  author       = {{Google DeepMind}},
  howpublished = {Google Blog},
  year         = {2025},
  url          = {https://blog.google/products-and-platforms/products/gemini/gemini-3/#responsible-development},
}

@misc{Grok4,
  title        = {{Grok-4}},
  author       = {{xAI}},
  howpublished = {xAI News},
  year         = {2025},
  url          = {https://x.ai/news/grok-4},
}

@article{Qwen3,
  title   = {{Qwen3 Technical Report}},
  author  = {{Qwen Team}},
  journal = {arXiv preprint arXiv:2505.09388},
  year    = {2025},
  url     = {https://qwen.ai/blog?id=qwen3},
}

@article{InternS1Pro,
  title   = {{Intern-S1-Pro: Scientific Multimodal Foundation Model at Trillion Scale}},
  author  = {{InternLM Team}},
  journal = {arXiv preprint arXiv:2603.25040},
  year    = {2026},
  url     = {https://arxiv.org/abs/2603.25040},
}

@Article{EuclidCollaboration2025b,
  author        = {{Euclid Collaboration} and Stevens, G. and Fotopoulou, S. and Bremer, M. N. and Zatarain, T. Matamoro and Jahnke, K. and Margalef-Bentabol, B. and Huertas-Company, M. and Smith, M. J. and Walmsley, M. and Salvato, M. and Mezcua, M. and Paulino-Afonso, A. and Siudek, M. and Talia, M. and Ricci, F. and Roster, W. and Aghanim, N. and Altieri, B. and Andreon, S. and Aussel, H. and Baccigalupi, C. and Baldi, M. and Bardelli, S. and Battaglia, P. and Biviano, A. and Bonchi, A. and Branchini, E. and Brescia, M. and Brinchmann, J. and Camera, S. and Cañas-Herrera, G. and Capobianco, V. and Carbone, C. and Carretero, J. and Castellano, M. and Castignani, G. and Cavuoti, S. and Chambers, K. C. and Cimatti, A. and Colodro-Conde, C. and Congedo, G. and Conselice, C. J. and Conversi, L. and Copin, Y. and Costille, A. and Courbin, F. and Courtois, H. M. and Cropper, M. and Da Silva, A. and Degaudenzi, H. and De Lucia, G. and Dolding, C. and Dole, H. and Douspis, M. and Dubath, F. and Dupac, X. and Dusini, S. and Escoffier, S. and Farina, M. and Ferriol, S. and George, K. and Giocoli, C. and Granett, B. R. and Grazian, A. and Grupp, F. and Haugan, S. V. H. and Hook, I. M. and Hormuth, F. and Hornstrup, A. and Hudelot, P. and Jhabvala, M. and Keihänen, E. and Kermiche, S. and Kiessling, A. and Kilbinger, M. and Kubik, B. and Kümmel, M. and Kurki-Suonio, H. and Boulc'h, Q. Le and Brun, A. M. C. Le and Mignant, D. Le and Lilje, P. B. and Lindholm, V. and Lloro, I. and Mainetti, G. and Maino, D. and Maiorano, E. and Marggraf, O. and Martinelli, M. and Martinet, N. and Marulli, F. and Massey, R. and Maurogordato, S. and McCracken, H. J. and Medinaceli, E. and Mei, S. and Melchior, M. and Meneghetti, M. and Merlin, E. and Meylan, G. and Mora, A. and Moresco, M. and Moscardini, L. and Nakajima, R. and Neissner, C. and Niemi, S. -M. and Padilla, C. and Paltani, S. and Pasian, F. and Pedersen, K. and Percival, W. J. and Pettorino, V. and Polenta, G. and Poncet, M. and Popa, L. A. and Pozzetti, L. and Raison, F. and Rebolo, R. and Renzi, A. and Rhodes, J. and Riccio, G. and Romelli, E. and Roncarelli, M. and Saglia, R. and Sánchez, A. G. and Sapone, D. and Schewtschenko, J. A. and Schirmer, M. and Schneider, P. and Schrabback, T. and Secroun, A. and Serrano, S. and Simon, P. and Sirignano, C. and Sirri, G. and Skottfelt, J. and Stanco, L. and Steinwagner, J. and Tallada-Crespí, P. and Taylor, A. N. and Tereno, I. and Toft, S. and Toledo-Moreo, R. and Torradeflot, F. and Tutusaus, I. and Valenziano, L. and Valiviita, J. and Vassallo, T. and Kleijn, G. Verdoes and Veropalumbo, A. and Wang, Y. and Weller, J. and Zacchei, A. and Zamorani, G. and Zerbi, F. M. and Zinchenko, I. A. and Zucca, E. and Allevato, V. and Ballardini, M. and Bolzonella, M. and Bozzo, E. and Burigana, C. and Cabanac, R. and Cappi, A. and Vigo, J. A. Escartin and Gabarra, L. and Hartley, W. G. and Martín-Fleitas, J. and Matthew, S. and Metcalf, R. B. and Pezzotta, A. and Pöntinen, M. and Risso, I. and Scottez, V. and Sereno, M. and Tenti, M. and Wiesmann, M. and Akrami, Y. and Alvi, S. and Andika, I. T. and Anselmi, S. and Archidiacono, M. and Atrio-Barandela, F. and Bertacca, D. and Bethermin, M. and Bisigello, L. and Blanchard, A. and Blot, L. and Borgani, S. and Brown, M. L. and Bruton, S. and Calabro, A. and Caro, F. and Castro, T. and Cogato, F. and Davini, S. and Desprez, G. and Díaz-Sánchez, A. and Diaz, J. J. and Di Domizio, S. and Diego, J. M. and Duc, P. -A. and Enia, A. and Fang, Y. and Ferrari, A. G. and Finoguenov, A. and Fontana, A. and Franco, A. and García-Bellido, J. and Gasparetto, T. and Gautard, V. and Gaztanaga, E. and Giacomini, F. and Gianotti, F. and Guidi, M. and Gutierrez, C. M. and Hall, A. and Hemmati, S. and Hildebrandt, H. and Hjorth, J. and Kajava, J. J. E. and Kang, Y. and Kansal, V. and Karagiannis, D. and Kirkpatrick, C. C. and Kruk, S. and Legrand, L. and Lembo, M. and Lepori, F. and Leroy, G. and Lesgourgues, J. and Leuzzi, L. and Liaudat, T. I. and Macias-Perez, J. and Magliocchetti, M. and Mannucci, F. and Maoli, R. and Martins, C. J. A. P. and Maurin, L. and Miluzio, M. and Monaco, P. and Morgante, G. and Naidoo, K. and Navarro-Alsina, A. and Passalacqua, F. and Paterson, K. and Patrizii, L. and Pisani, A. and Potter, D. and Quai, S. and Radovich, M. and Rocci, P. -F. and Rodighiero, G. and Sacquegna, S. and Sahlén, M. and Sanders, D. B. and Sarpa, E. and Schneider, A. and Schultheis, M. and Sciotti, D. and Sellentin, E. and Shankar, F. and Smith, L. C. and Tanidis, K. and Testera, G. and Teyssier, R. and Tosi, S. and Troja, A. and Tucci, M. and Valieri, C. and Vergani, D. and Verza, G. and Walton, N. A.},
  journal       = {A\&A},
  title         = {Euclid Quick Data Release (Q1). Active galactic nuclei identification using diffusion-based inpainting of Euclid VIS images},
  year          = {2025},
  issn          = {1432-0746},
  month         = oct,
  archiveprefix = {arXiv},
  copyright     = {Creative Commons Attribution 4.0 International},
  creationdate  = {2025-03-21T01:00:36},
  doi           = {10.1051/0004-6361/202554612},
  eprint        = {2503.15321},
  fjournal      = {Astronomy &amp; Astrophysics},
  primaryclass  = {astro-ph.GA},
  publisher     = {EDP Sciences},
}

@article{malz2019photometric,
  title={The photometric lsst astronomical time-series classification challenge plasticc: Selection of a performance metric for classification probabilities balancing diverse science goals},
  author={Malz, AI and Hlo{\v{z}}ek, Ren{\'e}e and Allam Jr, Tarek and Bahmanyar, Anita and Biswas, Rahul and Dai, Mi and Galbany, Llu{\'\i}s and Ishida, EEO and Jha, SW and Jones, DO and others},
  journal={The Astronomical Journal},
  volume={158},
  number={5},
  pages={171},
  year={2019},
  publisher={The American Astronomical Society}
}

@article{Asmus2020,
  author        = {Asmus, D. and Greenwell, C. L. and Gandhi, P. and Boorman, P. G. and Aird, J. and Alexander, D. M. and Assef, R. J. and Baldi, R. D. and Davies, R. I. and H{\"o}nig, S. F. and Ricci, C. and Rosario, D. J. and Salvato, M. and Shankar, F. and Stern, D.},
  title         = {Local {AGN} survey ({LASr}): {I}. {Galaxy} sample, infrared colour selection, and predictions for {AGN} within 100 {Mpc}},
  journal       = {Monthly Notices of the Royal Astronomical Society},
  volume        = {494},
  number        = {2},
  pages         = {1784--1816},
  year          = {2020},
  month         = may,
  doi           = {10.1093/mnras/staa766}
}

@article{Langis2026,
  author        = {Langis, D. A. and Liodakis, I. and Koljonen, K. I. I. and Paggi, A. and Globus, N. and Wyrzykowski, L. and Miko{\l}ajczyk, P. J. and Kotysz, K. and Zieli{\'n}ski, P. and Ihanec, N. and Ding, J. and Morshed, D. and Torres, Z.},
  title         = {Repeating flares, {X}-ray outbursts and delayed infrared emission: A comprehensive compilation of optical tidal disruption events: {TDECat}},
  journal       = {Astronomy \& Astrophysics},
  volume        = {707},
  pages         = {A171},
  year          = {2026},
  month         = mar,
  doi           = {10.1051/0004-6361/202553958}
}

@article{Scaife2021,
  author        = {Scaife, Anna M. M. and Porter, Florence},
  title         = {{Fanaroff--Riley} classification of radio galaxies using group-equivariant convolutional neural networks},
  journal       = {Monthly Notices of the Royal Astronomical Society},
  volume        = {503},
  number        = {2},
  pages         = {2369--2379},
  year          = {2021},
  month         = may,
  doi           = {10.1093/mnras/stab530}
}

\clearpage
\appendix
\section{Supplementary Materials}
\label{sec:supp}

\setcounter{table}{0}
\renewcommand{\thetable}{S\arabic{table}}

\setcounter{figure}{0}
\renewcommand{\thefigure}{S\arabic{figure}}

\noindent The supplementary material is organized as follows:
\begin{itemize}[leftmargin=*,nosep]
    \item Appendix~\ref{sec:supp-curation}: \emph{Detailed Dataset Curation}. Per-task curation procedures, sample selection, preprocessing, and de-identification.
    \item Appendix~\ref{sec:supp-reasoning}: \emph{Reasoning Case Studies and Expert Audit}. Verbatim reasoning excerpts, side-by-side comparison across models on Task~5, and quantitative expert evaluation of reasoning validity.
    \item Appendix~\ref{sec:supp-perf-details}: \emph{Detailed Per-Task Performance Analysis}. Per-task error analysis, class-wise failure modes, and confusion-matrix highlights.
    \item Appendix~\ref{sec:supp-stats}: \emph{Statistical Significance and Pairwise Model Comparisons}. Class-stratified bootstrap confidence intervals and paired randomization tests with Holm correction.
    \item Appendix~\ref{sec:supp-tables}: \emph{Supplementary Tables}. Per-model baseline accuracy tables and cross-model mechanistic ablation tables.
    \item Appendix~\ref{sec:supp-cm}: \emph{Confusion Matrices}. Row-normalized confusion matrices for every task and model.
    \item Appendix~\ref{sec:supp-cm-ablation}: \emph{Prompt Ablation Confusion Matrices}. Per-condition confusion matrices for the prompt-guidance ablation on Tasks~1 and~4.
    \item Appendix~\ref{sec:supp-prompts}: \emph{Guided Baseline Prompts}. Full prompt templates for every task and ablation condition.
\end{itemize}


\subsection{Detailed Dataset Curation}
\label{sec:supp-curation}

This section provides the additional dataset details that complement the per-task descriptions in Section~\ref{sec:methods} of the main paper. The five tasks are constructed entirely from publicly released astronomical surveys and catalogs to ensure reproducibility, with all class labels derived from spectroscopic confirmation or expert identification.

\paragraph{Task~1 (HSC-SSP optical imaging).}
Inputs are $g/r/i$-band RGB composites generated with \texttt{trilogy}~\cite{Coe2012} from HSC-SSP PDR3 cutouts~\cite{Aihara2017,Aihara2022}, centered on the galaxy nucleus and spanning $10\arcsec\times10\arcsec$. Sample: 300 spectroscopically confirmed Type-1 AGN~\cite{Li2021} with bright nuclear point sources, plus 257 inactive galaxies~\cite{Simard2011}. The sample is chosen to span a wide distribution of host-galaxy properties (effective radius and AGN-to-host/bulge-to-total ratio) so that decision boundaries are not driven by an idiosyncratic sub-population.

\paragraph{Task~2 (MiraBest radio morphology).}
The parent sample is the MiraBest dataset~\cite{Porter2023}, restricted to confidently classified FRI/FRII sources (excluding hybrid and uncertain entries) and yielding 397 FRI and 436 FRII. Each source has matched FIRST~\cite{Becker1995} and NVSS~\cite{Condon1998} radio maps; evaluating the identical sources at the two survey resolutions enables a direct measurement of how spatial detail affects morphological classification, which is otherwise confounded by sample composition.

\paragraph{Task~3 (AKARI NEP multi-wavelength SEDs).}
Inputs combine HSC ($g/r/i/z/y$), Euclid ($Y/J/H$), AKARI NEP~\cite{Kim2020} (N2, N3, N4, S7, S9W, S11, L15, L18, L24), and WISE (W1 to W4) photometry, cross-matched with the DESI DR1 AGN/QSO Value-Added Catalog. As noted in Section~\ref{sec:methods}, the candidate pool is bounded by the requirement of simultaneous panchromatic coverage across all four facilities and the intrinsic rarity of luminous local AGNs~\cite{Asmus2020}. After photometric-completeness pre-screening (2{,}142 from 4{,}842 candidates), we draw 44 spectroscopically confirmed Type-1 AGN, 40 Type-2 AGN identified by both NII-BPT~\cite{Baldwin1981} and SII-BPT~\cite{Veilleux1987} diagnostics, and 84 inactive galaxies (with no trace of nuclear activity). Final sample spans $0.04 < z < 0.45$. Each input is a $\nu f_\nu$ vs.\ wavelength plot (log-log) with both rest-frame and observed-frame axes, plus the spectroscopic redshift.

\paragraph{Task~4 (PLAsTiCC light curves).}
\label{sec:methods-task4}
Sample: 142 objects (30 AGN, 30 SNIa, 30 TDE, 30 RRL, 22 Mira) drawn from the PLAsTiCC simulated LSST-like dataset~\cite{plasticc2019}. To avoid severe class imbalance from distorting multi-class metrics, classes are balanced near 30 each. As described in Section~\ref{sec:methods}, the cohort size is constrained by catalog availability for well-sampled Miras ($N=22$) and the intrinsic astrophysical rarity of transients such as TDEs~\cite{Langis2026}. Each light curve presents flux-vs.-time across six LSST-like bands ($u/g/r/i/z/y$); PLAsTiCC supplies realistic cadence, noise, and detection thresholds.

\paragraph{Task~5 (DESI DR1 optical spectra).}
We curate 700 optical spectra from DESI~\cite{DESICollaboration2025}, organized into seven ground-truth sub-groups (A, B, C1, C2, C3, C4, D). Q1 (all 700) asks whether both H$\alpha$ ($\lambda$6563\,\AA) and H$\beta$ ($\lambda$4861\,\AA) emission lines are present; the Q1 negative class includes spectra in which one or both Balmer lines are absent because the source is at sufficiently high redshift that the lines fall outside the observed window, or because the source intrinsically lacks these emission features. Q2 (500, groups C1-C4 and D) asks whether the source is a Broad-Line AGN. Q3 (400, groups C1-C4) asks for the four-class BPT classification (Star-Forming, Composite, Seyfert, LINER), with labels derived from the Baldwin-Phillips-Terlevich diagnostic~\cite{Baldwin1981}. Redshifts and emission-line measurements come from the DESI \texttt{FastSpecFit} pipeline~\cite{Moustakas2023}; spectra are smoothed with a Gaussian kernel ($\sigma=5$ pixels) before display.

\paragraph{Data de-identification and pretraining contamination mitigation.}
To prevent models from relying on rote memorization of published catalogs, all observational inputs in \OURBENCH{} were strictly de-identified. Celestial coordinates ($\alpha, \delta$), survey IDs (e.g., SDSS SpecObjID, DESI Target ID, HSC Object ID), and catalog names were stripped from headers, tables, and figure labels. Models must evaluate observational data features directly. While public surveys may have appeared in the internet pretraining datasets of commercial foundation models, de-identification ensures that tasks evaluate multi-modal physical interpretation rather than database query retrieval.


\clearpage
\subsection{Reasoning Case Studies}
\label{sec:supp-reasoning}

This section expands on the qualitative reasoning analysis presented in the Discussion (Figure~\ref{fig:reasoning} of the main paper), providing the verbatim reasoning excerpts and the full side-by-side comparison across frontier models on Task~5 (Spectral Interpretation).

\paragraph{Autonomous reasoning by Gemini 3 Pro.}
The reasoning quality analysis on Task~5 reveals both impressive capabilities in frontier models and prevalent flaws in their logical processes. Gemini 3 Pro demonstrates impressive autonomous reasoning grounded in astronomical knowledge. For emission-line detection (Q1), the model correctly identifies the spectrum as a high-redshift quasar ($z \approx 2.25$) by recognizing Ly$\alpha$ ($\sim$3{,}950\,\AA) and C~IV ($\sim$5{,}000\,\AA) features. Notably, these lines were not provided in the prompt, indicating the model's ability to flexibly retrieve internal domain knowledge to determine the redshift and correctly infer that H$\alpha$ and H$\beta$ are redshifted out of the observational optical window. In the broad-line assessment (Q2), despite the absence of explicit instructions, the model autonomously contrasts the broad Balmer profiles (H$\alpha$ at $\sim$7{,}150\,\AA, H$\beta$ at $\sim$5{,}300\,\AA) with the narrow [O~III] forbidden lines ($\sim$5{,}460\,\AA) to confirm the high-velocity Broad Line Region. Furthermore, for the complex BPT classification (Q3), Gemini provides remarkably accurate quantitative emission-line flux estimates and correctly positions the galaxy on the diagnostic diagram, all without external measurement tools or explicit formulas in the prompt.

\paragraph{Fragile or flawed logic in other models.}
In contrast, other models exhibit fragile or flawed logic even when their final predictions are correct. For the Q1 example, Intern-S1 and GPT arrive at the correct binary answer but rely on a logically unrigorous assumption of a low-redshift source: they focus narrowly on a single feature, incorrectly assigning the $\sim$5{,}000\,\AA\ peak to H$\beta$, and deduce the answer merely from the absence of H$\alpha$. Claude and Grok demonstrate a slightly more comprehensive visual examination by successfully identifying multiple emission peaks. However, they ultimately fail the classification. Their wavelength localization is imprecise, leading them to erroneously misidentify C~III] as H$\alpha$. This assignment forces them to ignore the physical constraint that H$\alpha$ is typically brighter than H$\beta$, which directly contradicts the observed relative peak intensities in the spectrum. Fundamentally, their failure stems from an inability to extrapolate beyond the prompt; because they fail to spontaneously retrieve and associate the peaks with unprompted ultraviolet features like Ly$\alpha$ and C~IV, they forcibly map the observed spectrum onto the low-redshift emission lines queried in the prompt.

\paragraph{Quantitative expert evaluation of reasoning validity.}
To quantify the prevalence of the ``right-answer-wrong-reason'' phenomenon beyond individual case studies, an astrophysics domain expert evaluated a randomly sampled cohort of correctly classified instances from Task~5 (Optical Spectroscopy) across all six models ($N=200$ total evaluations across subtasks Q1, Q2, and Q3). For each instance, the expert examined the model's generated \texttt{"reason"} string alongside the corresponding ground-truth spectrum to determine whether the stated scientific logic was physically sound, consistent with the observational data, and causally responsible for the correct decision.

Across all evaluated models, \textbf{an average of 26\% of correct classifications were accompanied by physically invalid, contradictory, or unprincipled reasoning chains}. The primary failure modes identified by the expert include:
\begin{enumerate}[leftmargin=*, nosep]
    \item \emph{Imprecise wavelength localization:} In Task~5 Q1, models frequently latch onto noise spikes or unrelated continuum fluctuations 200--500\,\AA\ away from the rest-frame Balmer positions, misidentifying them as H$\alpha$ or H$\beta$ while still outputting the correct binary detection decision.
    \item \emph{Inverted Balmer decrements and intensity ratios:} Models frequently assert the presence of H$\beta$ at significantly higher amplitude than H$\alpha$ in unreddened star-forming systems, defying fundamental Case~B recombination physics without recognizing dust attenuation or instrumental artifacts.
    \item \emph{Confirmatory line hallucination on high-redshift spectra:} When evaluated on high-redshift quasars ($z > 1.5$), several models forcibly map observed ultraviolet features (such as C~IV $\lambda$1549 or C~III] $\lambda$1909) onto the queried low-redshift Balmer lines (H$\alpha$ and H$\beta$), failing to account for cosmological redshifting yet serendipitously arriving at the correct classification through flawed premises.
    \item \emph{Erroneous forbidden-line physics:} On Task~5 Q2 and Q3, models frequently conflate narrow forbidden lines ([O~III] $\lambda$5007, [N~II] $\lambda$6584) with broad permitted Balmer components, asserting broad-line emission in strictly narrow-line Seyfert~2 galaxies while arriving at the correct general category by chance.
\end{enumerate}
Among individual models, Gemini~3~Pro demonstrated the lowest rate of reasoning invalidity ($\approx 14\%$), followed by Qwen3-235B ($\approx 24\%$), Intern-S1-Pro ($\approx 25\%$), GPT-5.2 ($\approx 28\%$), Claude Opus 4.5 ($\approx 31\%$), and Grok-4 ($\approx 34\%$). These quantitative results demonstrate that headline classification accuracy can be deceptive in scientific domains, where ungrounded heuristics frequently masquerade as legitimate physical understanding.

\paragraph{Implications.}
These findings underscore the critical importance of evaluating the logical validity of the reasoning chain alongside accuracy. A model that yields a correct classification but cites incorrect spectral features provides a false sense of understanding that could mislead downstream scientific pipelines. The reasoning failures are consistent with our broader findings: models can often latch onto phenomenologically plausible local cues without deploying the physical knowledge required to justify those cues rigorously. While models like Gemini showcase the potential for physics-informed visual reasoning, the inconsistency and hallucination risks across frontier VLMs highlight that reasoning validation is essential for trustworthy scientific deployment.


\clearpage
\subsection{Detailed Per-Task Performance Analysis}
\label{sec:supp-perf-details}

This section expands on the per-task results summarized in Figure~2 of the main paper. Per-model accuracy tables and confusion matrices follow in the next subsections.

\paragraph{Task~1: QSO Host Galaxy Classification.}
Gemini~3~Pro achieves the highest accuracy at 74.5\% (95\% CI [0.709, 0.781]), followed by GPT-5.2 at 65.2\% [0.612, 0.691] and Grok-4 at 63.6\% [0.596, 0.675]. These three models substantially outperform Intern-S1-Pro (58.2\%), while Claude Opus 4.5 and Qwen3-235B both achieve only 46.1\%. Error analysis reveals the primary capability gap lies in point spread function (PSF) sensitivity. Gemini~3~Pro is the most balanced (76\% AGN, 73\% Galaxy specificity). GPT-5.2 maintains high galaxy specificity (84\%) but identifies only 49\% of AGN. Grok-4 achieves decent AGN recall (72\%) but at the cost of poor galaxy specificity (54\%). Intern-S1-Pro massively over-predicts PSFs (96\% AGN recall, 14\% galaxy specificity). Claude Opus 4.5 and Qwen3-235B classify all sources as galaxies (0\% AGN recall).

\paragraph{Task~2: Radio Galaxy Morphology.}
On the higher-resolution MiraBest\_F images, Gemini~3~Pro is best at 84.6\%, substantially outperforming Grok-4 (70.5\%), GPT-5.2 (66.1\%), Intern-S1-Pro (65.1\%), Qwen3-235B (52.3\%), and Claude Opus 4.5 (47.2\%). Performance degrades sharply on the lower-resolution MiraBest\_N images: Gemini~3~Pro 55.3\%, Qwen3-235B 52.3\%, GPT-5.2 50.1\%, Grok-4 49.7\%, Intern-S1-Pro 48.5\%, Claude Opus 4.5 46.6\%. The collapse is consistent with NVSS's poorer spatial resolution blurring the brightness structure that distinguishes core-brightened FRI from edge-brightened FRII systems.

\paragraph{Task~3: SED Classification.}
Gemini~3~Pro achieves the highest three-class accuracy at 57.7\%, followed by GPT-5.2 (54.2\%) and Claude Opus 4.5 (53.6\%). However, these scores are inflated by a severe conservative bias: top-tier models reach $>90\%$ recall on Galaxies but catastrophically fail to identify AGN, misclassifying most as normal galaxies. Conversely, Grok-4 and Intern-S1-Pro overpredict AGN while failing on galaxies (e.g., Grok-4 has only 14\% Galaxy recall). Even on simplified binary AGN-vs-Galaxy classification, Gemini~3~Pro reaches only 64.3\% accuracy with 32\% AGN recall (dropping to 18\% for GPT-5.2). The pervasive failures suggest a fundamental inability to extract subtle physical accretion signatures (optical continuum shape, mid-infrared dust excesses) directly from visualized SED plots.

\paragraph{Task~4: Light Curve Classification.}
Gemini~3~Pro achieves the highest accuracy at 61.3\% [0.535, 0.690], well above GPT-5.2 (38.7\%), Grok-4 (31.0\%), Intern-S1-Pro (31.0\%), Claude Opus 4.5 (21.1\%), and Qwen3-235B (21.1\%). Class-wise patterns reveal distinct failure modes. Claude Opus 4.5 and Qwen3-235B suffer mode collapse, classifying all sources as AGN. Grok-4 and Intern-S1-Pro exhibit strong biases toward AGN or TDEs. While Gemini~3~Pro and GPT-5.2 demonstrate better class separation, they share common physical confusions: both frequently misclassify TDEs as SNIa (40\% Gemini, 87\% GPT-5.2), reflecting the difficulty in distinguishing two transients with similar rise/decline shapes. Short-period RR Lyrae are largely misclassified as AGN (67\% Gemini, 70\% GPT-5.2), likely because sparse periodic variations appear visually indistinguishable from stochastic AGN variability. Nevertheless, Gemini~3~Pro successfully isolates Mira variables (86\%) and SNIa (77\%).

\paragraph{Task~5: Spectral Interpretation.}
On Q1 (emission line detection), Claude Opus 4.5 leads at 82.0\% by maintaining 96\% specificity and 76\% recall. Grok-4's superficially high accuracy (75.4\%) masks a severe over-prediction bias (95\% recall, 27\% specificity). Qwen3-235B exhibits complete mode collapse (28.6\%), failing to detect any emission-line spectra. On Q2 (BLAGN identification), raw accuracy scores are misleading due to class imbalance (400 narrow-line vs.\ 100 broad-line). Although Claude, Qwen3, GPT-5.2, and Intern-S1-Pro all cluster around 80\% accuracy, their confusion matrices show catastrophic mode collapse: they fail to detect broad lines almost entirely (recall $\le 3\%$). Grok-4 essentially guesses randomly ($1{:}1$ confusion, 50.0\% accuracy). Gemini~3~Pro is the singular exception, achieving 87.4\% by genuinely identifying 42\% of true broad-line sources while maintaining 99\% specificity. On Q3 (BPT classification), Claude, Grok-4, Intern-S1-Pro, and Qwen3-235B all suffer extreme mode collapse, predicting Star-Forming for nearly all spectra (23\% to 28\% accuracy). GPT-5.2 (44.5\%) avoids pure mode collapse but exhibits distributed confusion among Seyfert/LINER/Star-Forming. Gemini~3~Pro achieves the best performance (55.8\%) by maintaining distinct representations for Star-Forming (96\% recall) and Seyfert (66\% recall), though it still struggles with Composite and LINER spectra. The progressive Q1$\to$Q2$\to$Q3 difficulty reveals that VLMs can identify localized signal spikes but largely fail at line-profile recognition and ratio-based diagnostics, with Gemini~3~Pro's stronger inductive biases for fine numerical detail being the main exception.


\clearpage
\subsection{Supplementary Tables}
\label{sec:supp-tables}

\begin{table}[h]
\centering
\caption{\textbf{Task 1: QSO Host Galaxy Classification.} Accuracy with 95\% bootstrap confidence intervals for binary AGN/Galaxy classification ($N=557$; 10{,}000 bootstrap iterations).}
\label{tab:supp-task1}
\begin{tabular}{lcc}
\toprule
Model & Accuracy & 95\% CI \\
\midrule
Gemini 3 Pro     & \textbf{0.745} & [0.709, 0.781] \\
GPT-5.2          & 0.652 & [0.612, 0.691] \\
Grok-4           & 0.636 & [0.596, 0.675] \\
Intern-S1-Pro    & 0.582 & [0.540, 0.623] \\ 
Claude Opus 4.5  & 0.461 & [0.420, 0.503] \\ 
Qwen3-235B       & 0.461 & [0.420, 0.503] \\
\bottomrule
\end{tabular}
\end{table}

\begin{table}[h]
\centering
\caption{\textbf{Task 2: Radio Galaxy Morphology.} Accuracy with 95\% bootstrap confidence intervals for Fanaroff-Riley Type I/II classification on two radio survey datasets: MiraBest\_F (FIRST survey, higher resolution) and MiraBest\_N (NVSS survey, lower resolution). $N=833$; 10{,}000 bootstrap iterations.}
\label{tab:supp-task2}
\begin{tabular}{lcccc}
\toprule
Model & MiraBest\_F & 95\% CI & MiraBest\_N & 95\% CI \\
\midrule
Gemini 3 Pro     & \textbf{0.846} & [0.821, 0.870] & \textbf{0.553} & [0.519, 0.587] \\ 
Grok-4           & 0.705 & [0.674, 0.735] & 0.497 & [0.463, 0.531] \\
GPT-5.2          & 0.661 & [0.630, 0.694] & 0.501 & [0.466, 0.534] \\
Intern-S1-Pro    & 0.651 & [0.617, 0.683] & 0.485 & [0.450, 0.519] \\
Qwen3-235B       & 0.523 & [0.490, 0.558] & 0.523 & [0.490, 0.558] \\
Claude Opus 4.5  & 0.472 & [0.438, 0.507] & 0.466 & [0.432, 0.499] \\
\bottomrule
\end{tabular}
\end{table}

\begin{table}[h]
\centering
\caption{\textbf{Task 3: SED Classification.} Accuracy with 95\% bootstrap confidence intervals for the three-class classification (Type-1 AGN, Type-2 AGN, Galaxy; $N = 168$; 10{,}000 bootstrap iterations). Per-class recall is reported for each source type: Galaxy ($N=84$), Type-1 AGN ($N=44$), and Type-2 AGN ($N=40$).}
\label{tab:supp-task3}
\begin{tabular}{lccccc}
\toprule
Model & Overall & 95\% CI & Galaxy & Type-1 AGN & Type-2 AGN \\
\midrule
Gemini 3 Pro     & \textbf{0.577} & [0.500, 0.649] & 0.964 & 0.091 & 0.300 \\
GPT-5.2          & 0.542 & [0.464, 0.619] & 0.988 & 0.000 & 0.200 \\
Claude Opus 4.5  & 0.536 & [0.458, 0.613] & 0.917 & 0.136 & 0.175 \\
Intern-S1-Pro    & 0.333 & [0.262, 0.405] & 0.250 & 0.273 & 0.575 \\
Grok-4           & 0.321 & [0.250, 0.393] & 0.143 & 0.432 & 0.575 \\
Qwen3-235B       & 0.262 & [0.196, 0.327] & 0.000 & 1.000 & 0.000 \\
\bottomrule
\end{tabular}
\end{table}

\begin{table}[h]
\centering
\caption{\textbf{Task 4: Light Curve Classification.} Accuracy with 95\% bootstrap confidence intervals for the five-class AGN/SNIa/TDE/RRL/Mira classification ($N=142$; 10{,}000 bootstrap iterations).}
\label{tab:supp-task4}
\begin{tabular}{lcc}
\toprule
Model & Accuracy & 95\% CI \\
\midrule
Gemini 3 Pro     & \textbf{0.613} & [0.535, 0.690] \\
GPT-5.2          & 0.387 & [0.310, 0.465] \\
Grok-4           & 0.310 & [0.239, 0.387] \\
Intern-S1-Pro    & 0.310 & [0.232, 0.387] \\
Claude Opus 4.5  & 0.211 & [0.148, 0.282] \\
Qwen3-235B       & 0.211 & [0.148, 0.282] \\
\bottomrule
\end{tabular}
\end{table}

\begin{table}[h]
\centering
\caption{\textbf{Task 5: Spectral Interpretation.} Accuracy with 95\% bootstrap confidence intervals for the three hierarchical spectral analysis questions (10{,}000 bootstrap iterations). Q1 ($N=700$): detection of both H$\alpha$ and H$\beta$ emission lines. Q2 ($N=500$): broad-line AGN identification. Q3 ($N=400$): BPT diagram classification into Star-Forming, Composite, Seyfert, or LINER.}
\label{tab:supp-task5}
\begin{tabular}{lcccccc}
\toprule
Model & Q1 & 95\% CI & Q2 & 95\% CI & Q3 & 95\% CI \\
\midrule
Claude Opus 4.5  & \textbf{0.820} & [0.791, 0.849] & 0.806 & [0.770, 0.840] & 0.278 & [0.235, 0.323] \\
GPT-5.2          & 0.756 & [0.723, 0.787] & 0.798 & [0.762, 0.834] & 0.445 & [0.395, 0.495] \\
Grok-4           & 0.754 & [0.723, 0.786] & 0.500 & [0.456, 0.546] & 0.238 & [0.198, 0.280] \\
Gemini 3 Pro     & 0.739 & [0.706, 0.771] & \textbf{0.874} & [0.844, 0.902] & \textbf{0.558} & [0.510, 0.608] \\
Intern-S1-Pro    & 0.664 & [0.629, 0.699] & 0.802 & [0.766, 0.836] & 0.260 & [0.218, 0.303] \\
Qwen3-235B       & 0.286 & [0.251, 0.319] & 0.800 & [0.764, 0.834] & 0.248 & [0.205, 0.290] \\
\bottomrule
\end{tabular}
\end{table}


\clearpage
\subsection{Statistical Significance and Pairwise Model Comparisons}
\label{sec:supp-stats}

To evaluate whether observed performance differences between models are statistically meaningful, particularly on tasks with constrained sample sizes or class imbalance, we implemented a rigorous non-parametric statistical testing protocol:
\begin{itemize}[leftmargin=*, nosep]
    \item \textbf{Class-stratified bootstrap confidence intervals:} For each model and task, we compute 95\% confidence intervals using 10{,}000 stratified bootstrap resamples (seed 42), preserving the exact empirical class proportions across iterations.
    \item \textbf{Paired randomization tests:} For every pair of models on each task, we conduct a two-sided paired randomization test ($10{,}000$ permutations) on instance-by-instance binary correctness vectors. This non-parametric test makes no distributional assumptions and accounts for instance-level correlations.
    \item \textbf{Holm-Bonferroni correction:} To strictly control the family-wise error rate across all pairwise comparisons within each task ($\binom{6}{2} = 15$ pairwise tests per task setting), we apply the Holm-Bonferroni step-down correction at an adjusted significance threshold of $\alpha = 0.05$.
\end{itemize}

Table~\ref{tab:supp-significance} summarizes the statistically significant pairwise differences across all five tasks following Holm correction. Key findings include:
\begin{itemize}[leftmargin=*, nosep]
    \item \textbf{Clear model dominance:} Gemini~3~Pro statistically significantly outperforms all other five models on Task~1 (HSC imaging, $p_{\text{adj}} < 0.01$), Task~2 FIRST (high-resolution radio, $p_{\text{adj}} < 0.01$), Task~4 (light curves, $p_{\text{adj}} < 0.01$), Task~5 Q2 (BLAGN, $p_{\text{adj}} < 0.01$), and Task~5 Q3 (BPT diagnostics, $p_{\text{adj}} < 0.01$). Claude Opus 4.5 statistically significantly outperforms all other models on Task~5 Q1 (Balmer line detection, $p_{\text{adj}} < 0.01$).
    \item \textbf{Statistically indistinguishable cohorts:} On Task~3 (SED classification), bootstrap confidence intervals overlap broadly (Gemini [0.500, 0.649], GPT-5.2 [0.464, 0.619], Claude Opus 4.5 [0.458, 0.613]), and no pairwise comparisons achieve statistical significance after Holm correction ($p_{\text{adj}} > 0.05$). Similarly, on Task~2 NVSS, low spatial resolution degrades all models into an overlapping regime (55.3\% to 46.6\%), where differences between most models are statistically indistinguishable.
\end{itemize}

\begin{table}[h]
\centering
\small
\caption{\textbf{Pairwise statistical significance across all tasks.} Statistically significant model comparisons from two-sided paired randomization tests ($10{,}000$ permutations) after Holm-Bonferroni multiple-testing correction ($\alpha = 0.05$). Tasks without significant pairwise differences are denoted by `None'.}
\label{tab:supp-significance}
\begin{tabular}{ll}
\toprule
Task / Setting & Statistically Significant Comparisons (Holm-corrected $p < 0.05$) \\
\midrule
Task~1 (HSC Imaging)        & Gemini~3~Pro $>$ GPT-5.2, Grok-4, Intern-S1-Pro, Claude Opus 4.5, Qwen3-235B \\
Task~2 (MiraBest FIRST)     & Gemini~3~Pro $>$ all models; Grok-4 $>$ GPT-5.2, Intern-S1, Qwen3, Claude \\
Task~2 (MiraBest NVSS)      & Gemini~3~Pro $>$ Claude Opus 4.5, Intern-S1-Pro \\
Task~3 (AKARI SED)          & None (all pairwise comparisons $p_{\text{adj}} > 0.05$; overlapping CIs) \\
Task~4 (PLAsTiCC Light Curves) & Gemini~3~Pro $>$ all models; GPT-5.2 $>$ Claude Opus 4.5, Qwen3-235B \\
Task~5 Q1 (Line Detection)  & Claude Opus 4.5 $>$ all models; Gemini~3~Pro, GPT-5.2, Grok-4 $>$ Qwen3-235B \\
Task~5 Q2 (BLAGN)           & Gemini~3~Pro $>$ all models \\
Task~5 Q3 (BPT Diagram)     & Gemini~3~Pro $>$ all models; GPT-5.2 $>$ Grok-4, Intern-S1, Qwen3, Claude \\
\bottomrule
\end{tabular}
\end{table}

\begin{table}[h]
\centering
\small
\caption{\textbf{Cross-model prompt grounding ablation.} Classification accuracy (\%) on Task~1 ($N=557$) and Task~4 ($N=142$) across three prompt conditions: Unguided, Physical (Guided), and Phenomenological for Gemini~3~Pro, GPT-5.2, and Claude Opus 4.5.}
\label{tab:supp-prompt-ablation}
\begin{tabular}{llccc}
\toprule
Task & Prompt Condition & Gemini~3~Pro & GPT-5.2 & Claude Opus 4.5 \\
\midrule
\multirow{3}{*}{Task~1 (Optical Imaging)} 
 & Unguided          & 64.1\% & 67.32\% & \textbf{50.63\%} \\
 & Physical (Guided) & \textbf{74.5\%} & \textbf{68.04\%} & 46.32\% \\
 & Phenomenological  & 69.7\% & 61.04\% & 48.83\% \\
\midrule
\multirow{3}{*}{Task~4 (Light Curves)} 
 & Unguided (Image)          & 52.8\% & 37.32\% & 33.10\% \\
 & Physical (Image)          & 61.3\% & 35.92\% & 30.28\% \\
 & Phenomenological (Image)  & \textbf{65.5\%} & \textbf{43.66\%} & \textbf{42.25\%} \\
\bottomrule
\end{tabular}
\end{table}

\begin{table}[h]
\centering
\small
\caption{\textbf{Cross-model representation ablation details.} Classification accuracy (\%) on Task~3 ($N=168$) and Task~4 ($N=142$) across image and tabular inputs for Gemini~3~Pro, GPT-5.2, and Claude Opus 4.5.}
\label{tab:supp-rep-ablation}
\begin{tabular}{llccc}
\toprule
Task & Representation Format & Gemini~3~Pro & GPT-5.2 & Claude Opus 4.5 \\
\midrule
\multirow{2}{*}{Task~3 (SED)} 
 & Image (Plot)              & \textbf{57.7\%} & 38.10\%          & \textbf{35.12\%} \\
 & Numerical Table (Ordered) & 51.8\%          & \textbf{57.74\%} & 25.60\% \\
\midrule
\multirow{3}{*}{Task~4 (Light Curves)} 
 & Image (Plot)              & 61.3\%          & 35.92\%          & 30.28\% \\
 & Numerical Table (Ordered) & \textbf{74.6\%} & \textbf{49.30\%} & \textbf{52.82\%} \\
 & Numerical Table (Shuffled) & 68.3\%         & 41.55\%          & 40.14\% \\
\bottomrule
\end{tabular}
\end{table}

\begin{table}[h]
\centering
\small
\caption{\textbf{Cross-model few-shot instruction ablation details.} Classification accuracy (\%) under zero-shot vs.\ few-shot visual exemplars across Gemini~3~Pro, GPT-5.2, and Claude Opus 4.5.}
\label{tab:supp-fewshot-ablation}
\begin{tabular}{lcccccc}
\toprule
\multirow{2}{*}{Setting} & \multicolumn{2}{c}{Gemini~3~Pro} & \multicolumn{2}{c}{GPT-5.2} & \multicolumn{2}{c}{Claude Opus 4.5} \\
\cmidrule(lr){2-3} \cmidrule(lr){4-5} \cmidrule(lr){6-7}
 & Zero-shot & Few-shot & Zero-shot & Few-shot & Zero-shot & Few-shot \\
\midrule
Task~2 MiraBest\_F & \textbf{84.6\%} & 83.1\% & 58.34\% & \textbf{59.66\%} & 39.14\% & \textbf{51.74\%} \\
Task~2 MiraBest\_N & \textbf{55.3\%} & 53.3\% & \textbf{52.10\%} & 50.66\% & 44.90\% & \textbf{47.90\%} \\
Task~5 Q1          & 73.9\% & \textbf{78.9\%} & 57.00\% & \textbf{72.57\%} & 39.57\% & \textbf{55.43\%} \\
Task~5 Q2          & \textbf{87.4\%} & 86.4\% & 81.40\% & \textbf{84.80\%} & 77.60\% & \textbf{81.56\%} \\
\bottomrule
\end{tabular}
\end{table}


\clearpage
\subsection{Confusion Matrices}
\label{sec:supp-cm}

\begin{figure}[h]
\centering
\includegraphics[width=\textwidth]{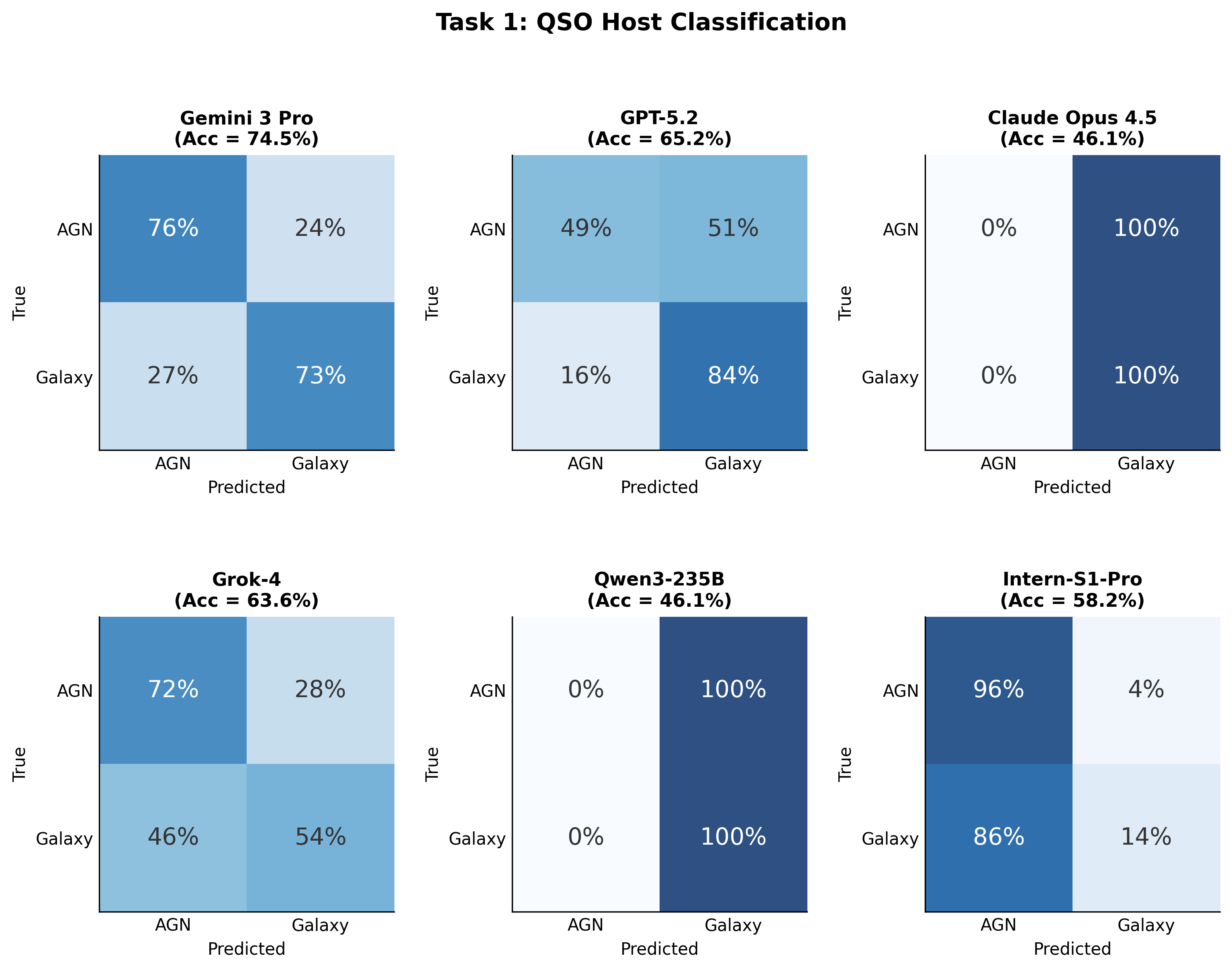}
\caption{\textbf{Task 1: QSO Host Galaxy Classification:Confusion Matrices.} Row-normalized confusion matrices for binary AGN/Galaxy classification across all six models.}
\label{fig:supp-cm-task1}
\end{figure}

\begin{figure}[h]
\centering
\includegraphics[width=\textwidth]{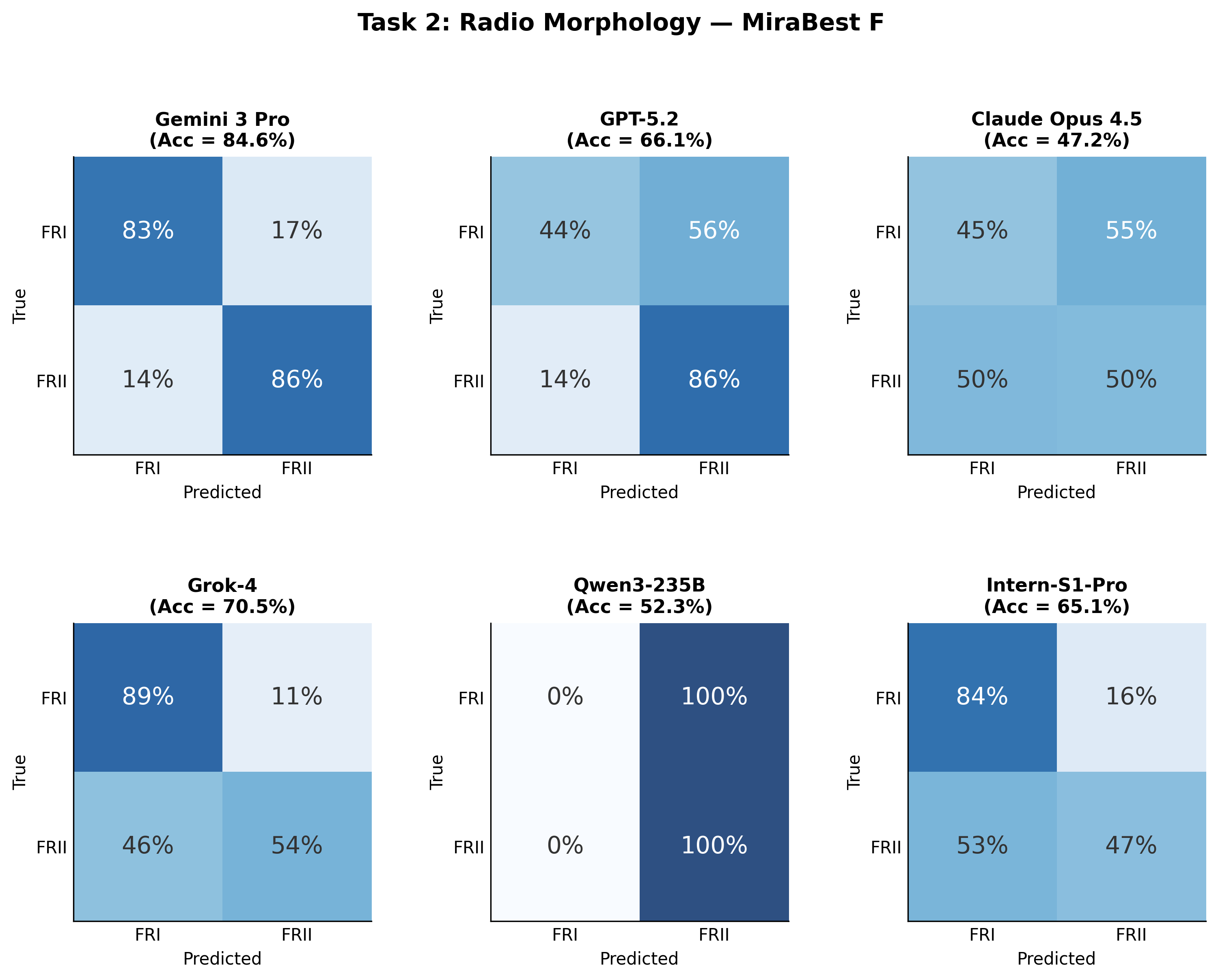}
\caption{\textbf{Task 2: Radio Galaxy Morphology:MiraBest F (FIRST).} Row-normalized confusion matrices for FRI/FRII classification on the higher-resolution FIRST survey data.}
\label{fig:supp-cm-task2f}
\end{figure}

\begin{figure}[h]
\centering
\includegraphics[width=\textwidth]{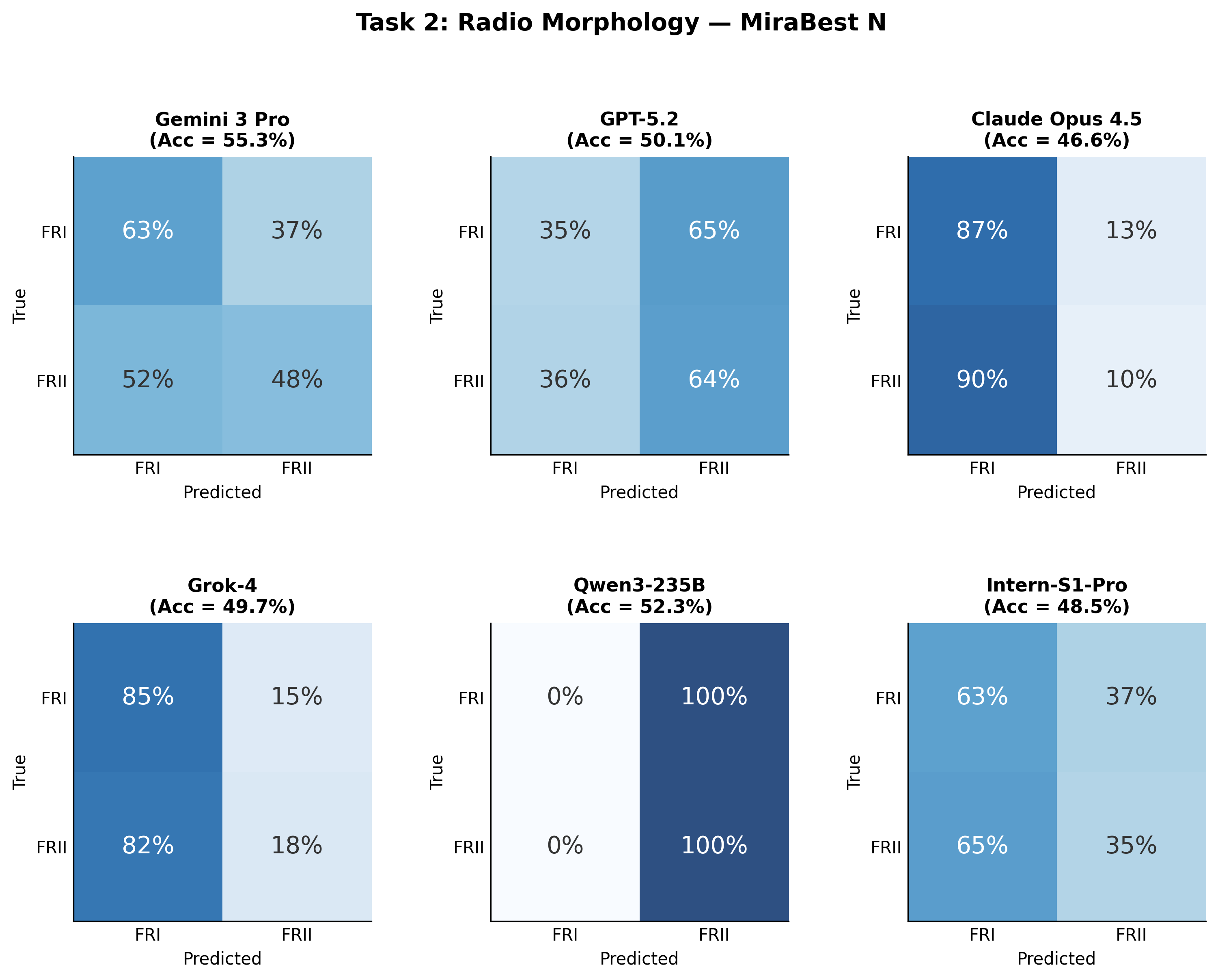}
\caption{\textbf{Task 2: Radio Galaxy Morphology:MiraBest N (NVSS).} Row-normalized confusion matrices for FRI/FRII classification on the lower-resolution NVSS survey data.}
\label{fig:supp-cm-task2n}
\end{figure}

\clearpage

\begin{figure}[h]
\centering
\includegraphics[width=\textwidth]{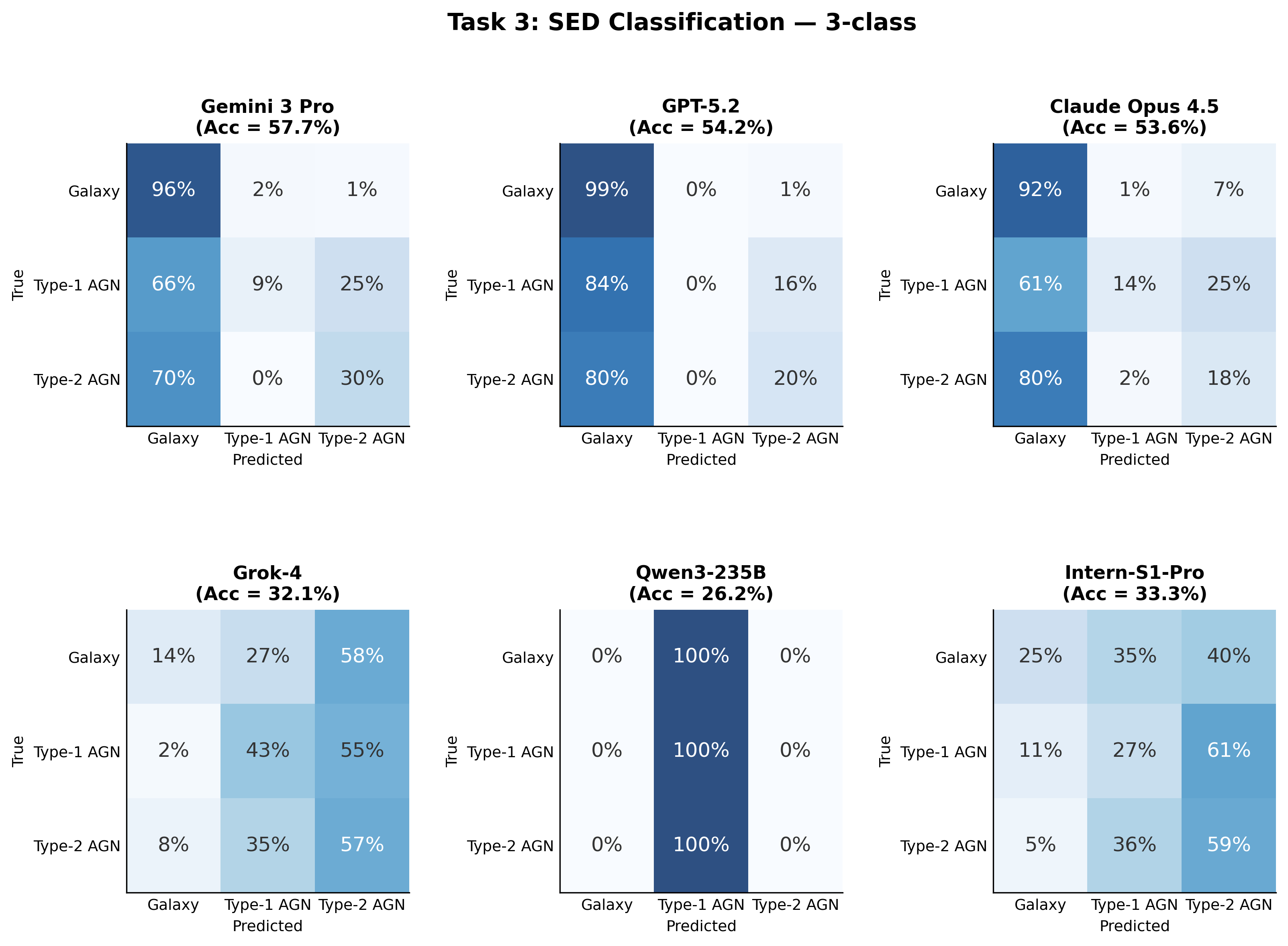}
\caption{\textbf{Task 3: SED Classification:3-class.} Row-normalized confusion matrices for Galaxy / Type-1 AGN / Type-2 AGN classification.}
\label{fig:supp-cm-task3-3class}
\end{figure}

\begin{figure}[h]
\centering
\includegraphics[width=\textwidth]{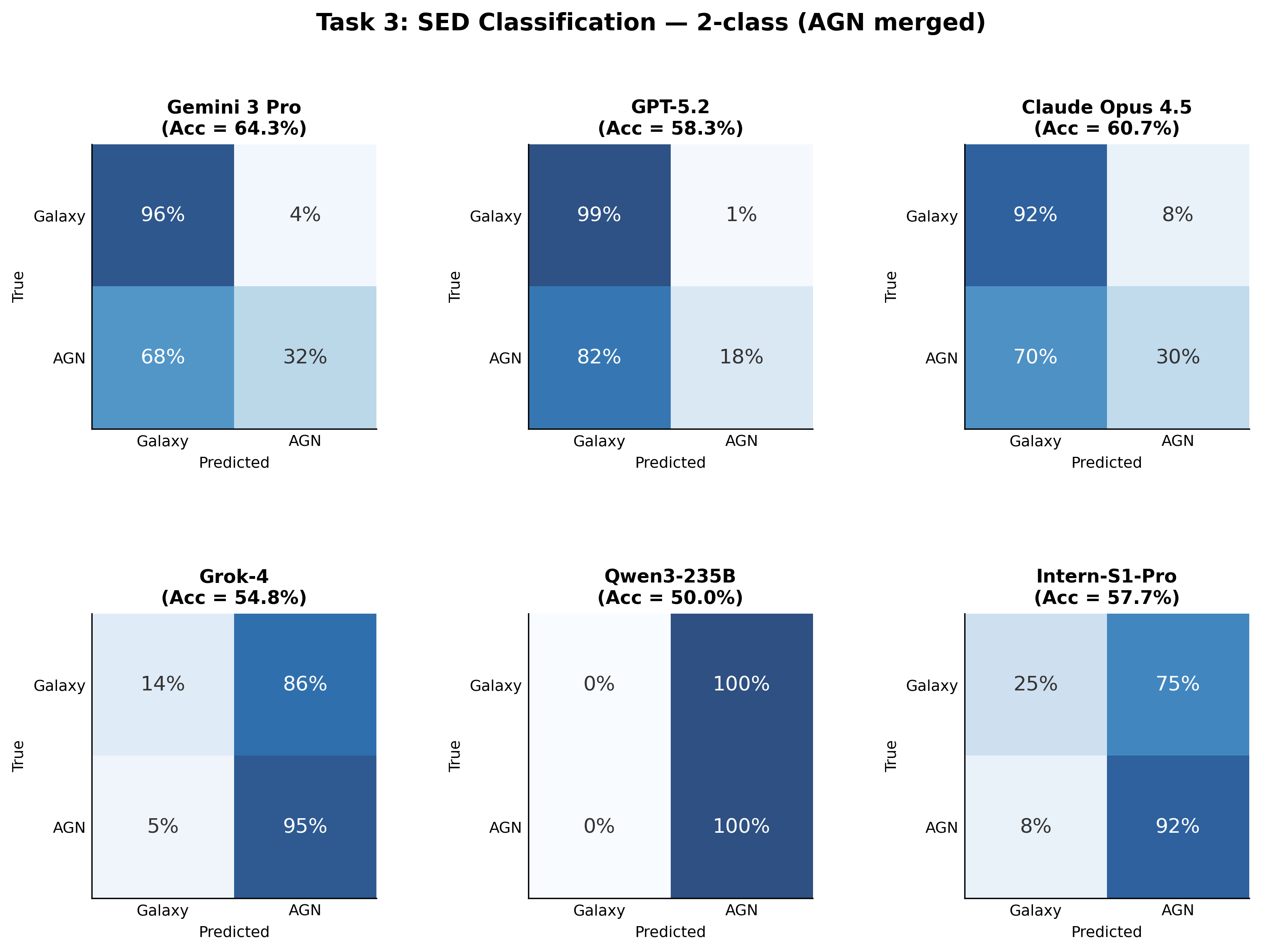}
\caption{\textbf{Task 3: SED Classification:2-class (AGN merged).} Row-normalized confusion matrices after merging Type-1 and Type-2 AGN into a single AGN class.}
\label{fig:supp-cm-task3-2class}
\end{figure}

\clearpage

\begin{figure}[h]
\centering
\includegraphics[width=\textwidth]{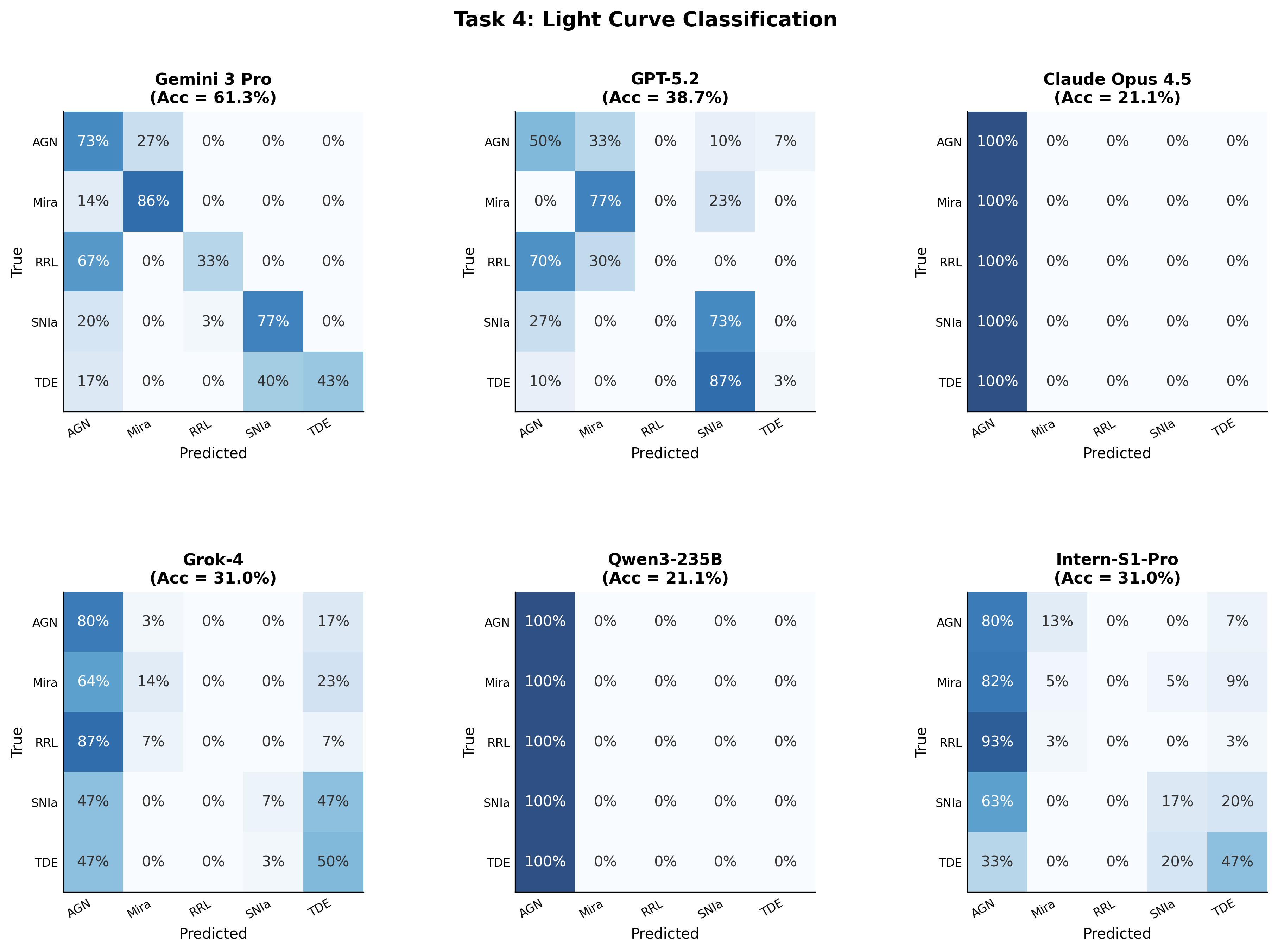}
\caption{\textbf{Task 4: Light Curve Classification:Confusion Matrices.} Row-normalized confusion matrices for the five-class AGN/Mira/RRL/SNIa/TDE classification.}
\label{fig:supp-cm-task4}
\end{figure}

\begin{figure}[h]
\centering
\includegraphics[width=\textwidth]{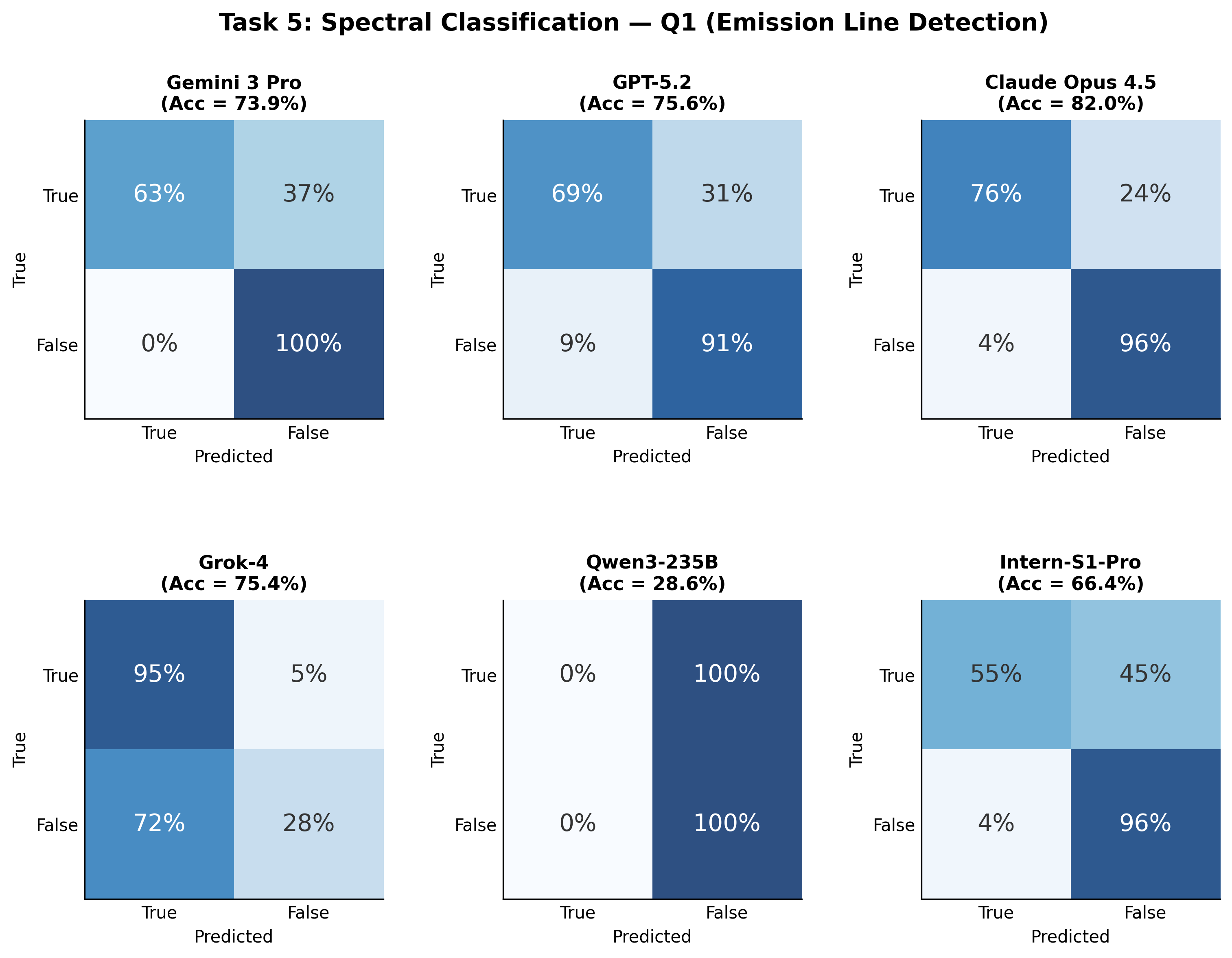}
\caption{\textbf{Task 5: Spectral Interpretation:Q1 (Emission Line Detection).} Row-normalized confusion matrices for binary True/False classification of H$\alpha$ and H$\beta$ emission line detection.}
\label{fig:supp-cm-task5q1}
\end{figure}

\clearpage

\begin{figure}[h]
\centering
\includegraphics[width=\textwidth]{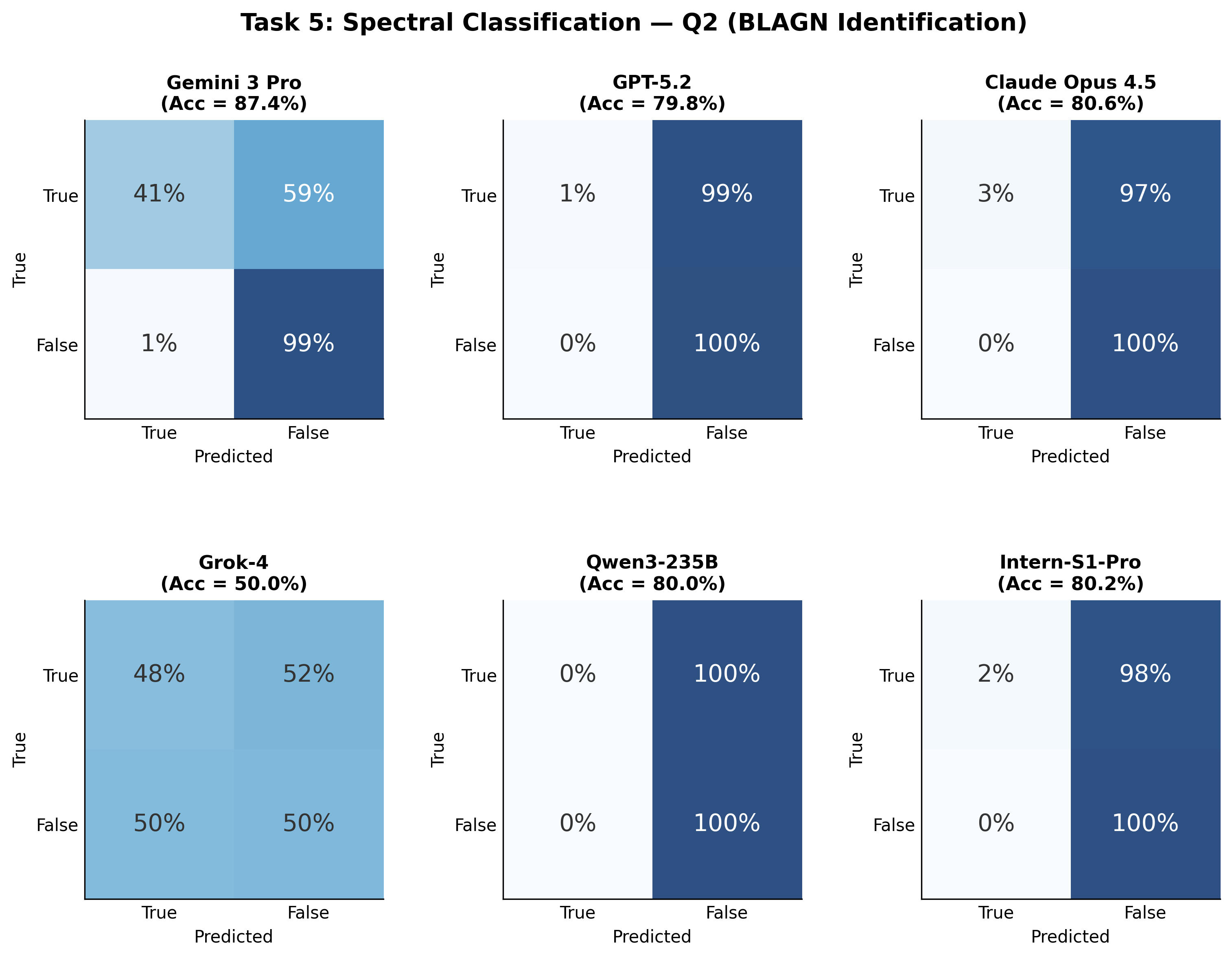}
\caption{\textbf{Task 5: Spectral Interpretation:Q2 (BLAGN Identification).} Row-normalized confusion matrices for binary True/False broad-line AGN identification.}
\label{fig:supp-cm-task5q2}
\end{figure}

\begin{figure}[h]
\centering
\includegraphics[width=\textwidth]{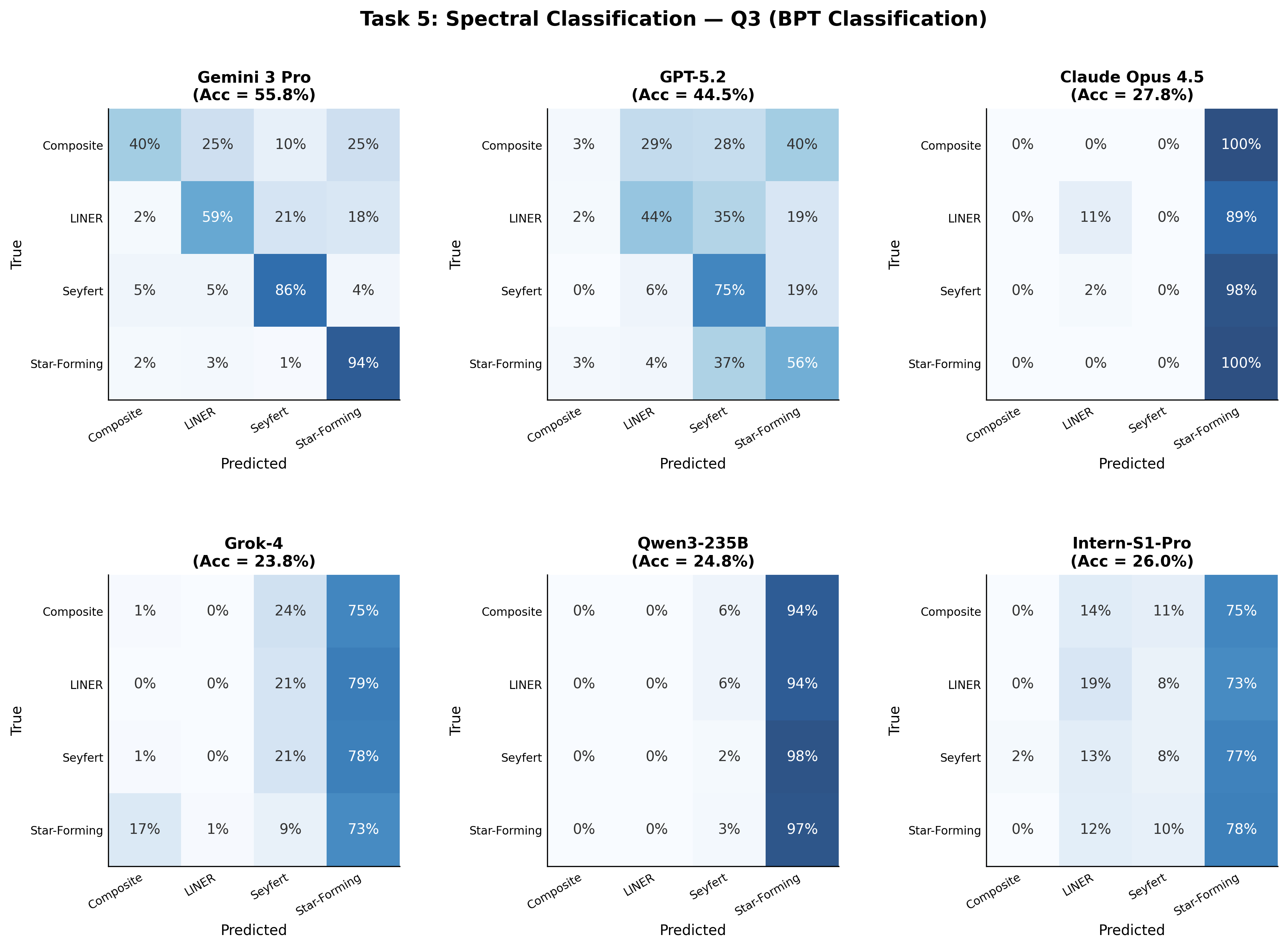}
\caption{\textbf{Task 5: Spectral Interpretation:Q3 (BPT Classification).} Row-normalized confusion matrices for the four-class BPT classification (Composite/LINER/Seyfert/Star-Forming).}
\label{fig:supp-cm-task5q3}
\end{figure}


\clearpage
\subsection{Prompt Ablation Confusion Matrices}
\label{sec:supp-cm-ablation}

\begin{figure}[h]
\centering
\includegraphics[width=\textwidth]{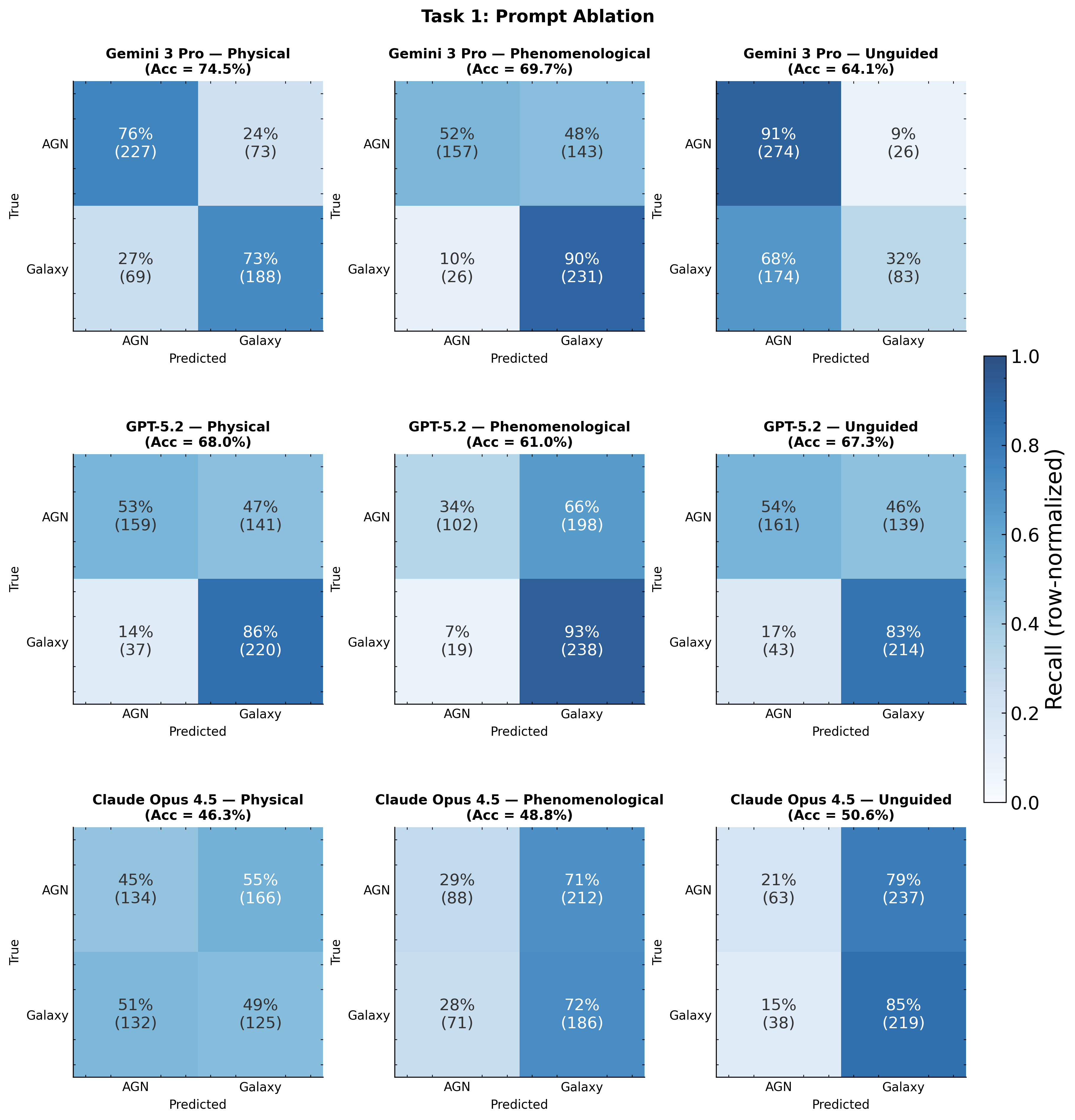}
\caption{\textbf{Task~1: Prompt Ablation across Frontier Models.} Row-normalized confusion matrices for binary AGN/Galaxy classification ($N=557$) under physical (left), phenomenological (center), and unguided (right) prompt conditions across Gemini~3~Pro (top row), GPT-5.2 (middle row), and Claude Opus~4.5 (bottom row). Physical prompting yields the most balanced decision boundaries for Gemini~3~Pro (76\% AGN, 73\% Galaxy) and GPT-5.2 (53\% AGN, 86\% Galaxy). Phenomenological prompting skews all models toward Galaxy predictions (90\%, 93\%, and 72\% Galaxy recall, respectively) by emphasizing diffuse host features over central point sources. Claude Opus~4.5 hovers near random performance (46.3--50.6\%), with its nominal unguided accuracy driven by a severe collapse toward Galaxy (85\% recall).}
\label{fig:supp-cm-task1-ablation}
\end{figure}

\begin{figure}[h]
\centering
\includegraphics[width=\textwidth]{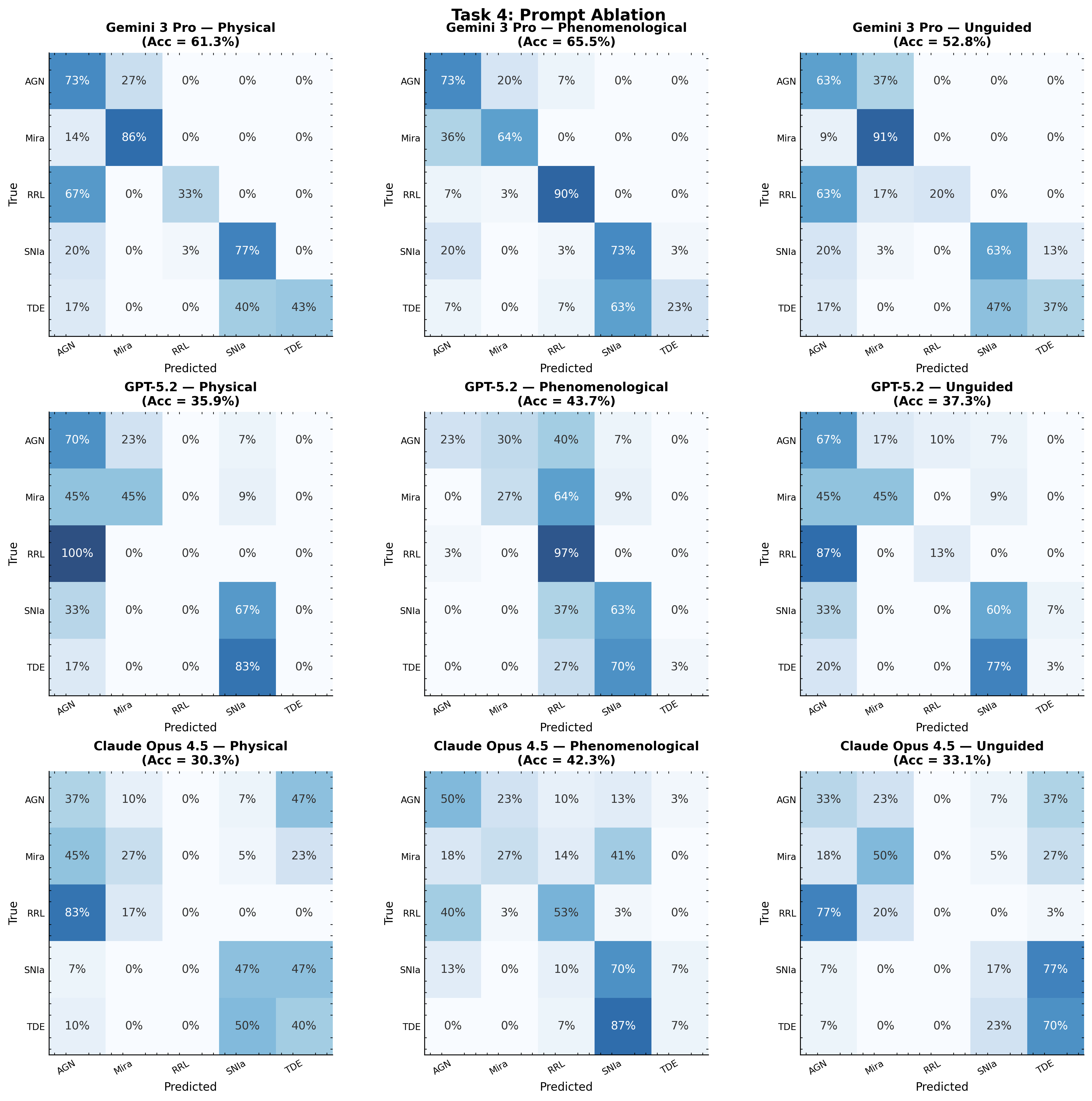}
\caption{\textbf{Task~4: Prompt Ablation across Frontier Models.} Row-normalized confusion matrices for five-class light curve classification ($N=142$) under physical (left), phenomenological (center), and unguided (right) prompt conditions across Gemini~3~Pro (top row), GPT-5.2 (middle row), and Claude Opus~4.5 (bottom row). Phenomenological guidance universally unblocks RR Lyrae (RRL) recognition across models by describing cadence-induced aliasing artifacts, yielding dramatic recall gains (Gemini: 33\%$\to$90\%; GPT-5.2: 0\%$\to$97\%; Claude: 0\%$\to$53\%). While Gemini~3~Pro preserves balanced boundaries on non-pulsating classes (AGN: 73\%, SNIa: 73\%, Mira: 64\%), lower-performing architectures exhibit heuristic over-correction (GPT-5.2 misclassifies 64\% of Miras and 40\% of AGNs as RRL) or severe mode collapse (Claude misclassifies 87\% of TDEs as SNIa).}
\label{fig:supp-cm-task4-ablation}
\end{figure}


\clearpage
\subsection{Guided Baseline Prompts}
\label{sec:supp-prompts}

\begin{prompt}
\label{prompt:task1}
\textbf{Prompt S1: Task 1: QSO Host Galaxy Classification (Physical/Guided).}\newline
\textbf{Task:} Classify this image as either AGN or Galaxy. \newline
\textbf{Instructions:}
Both categories are massive galaxies consisting of billions of stars spread over thousands of light-years. The distinction lies entirely in the physical mechanism dominating the very core:\newline
- Normal Galaxy: The light emission from the central region is purely the collective glow of the inner stellar populations, typically forming a central ``bulge.'' A bulge is a massive, spheroidal concentration of older stars. While the stellar density naturally increases and peaks at the very center, the emission remains fundamentally rooted in a thermal population distributed across a spatial volume spanning hundreds to thousands of light-years.\newline
- Type-1 AGN: In addition to the host galaxy's extended stellar components (like the disk and the central bulge), the exact geometric center hosts a supermassive black hole with an active accretion disk. This disk emits a staggering amount of radiative energy from a physically microscopic volume (often just light-days across). This results in a hyper-luminous, practically dimensionless energy source sitting directly on top of the extended stellar bulge.\newline
\newline
Examine the image, focusing on the interplay between the central emission and the extended host structure. Rely on your understanding of how these underlying physical realities would naturally manifest visually:\newline
- If your analysis indicates the \emph{presence} of an active accretion disk governing the core's emission, respond: AGN\newline
- If your analysis indicates the core is purely governed by distributed stellar emission (e.g., the bulge) without an overpowering active black hole component, respond: Galaxy\newline
\textbf{Output requirements:}\newline
- Respond with a JSON object in the following format: \{``answer'': ``'', ``reason'': ``''\}\newline
- The ``answer'' field must be either: AGN or Galaxy\newline
- The ``reason'' field should contain a brief explanation of your classification decision, explicitly linking the visual manifestation to the physical presence or absence of the central accretion disk versus a purely stellar bulge.\newline
- Do not include any text outside the JSON object
\end{prompt}

\begin{prompt}
\label{prompt:task1-phenom}
\textbf{Prompt S1b: Task 1: QSO Host Galaxy Classification (Phenomenological).}\newline
\textbf{Task:} Classify this image as either AGN or Galaxy. \newline
\textbf{Instructions:}
Examine the central region of this galaxy image in high detail. Follow these steps internally:\newline
1. Analyze core sharpness: determine whether the core is an unresolved point source (PSF-like) or a resolved, extended structure.\newline
2. Check for optical artifacts such as diffraction spikes or saturation bleeding.\newline
3. Evaluate the transition from the center to the disk: smooth and gradual (bulge-like) versus a sharp point superposed on extended light (AGN-like).\newline
- If the center is dominated by an unresolved point source consistent with a Type-1 AGN, respond: AGN\newline
- If the center is dominated by a resolved stellar bulge, respond: Galaxy\newline
\textbf{Output requirements:}\newline
- Respond with a JSON object in the following format: \{``answer'': ``'', ``reason'': ``''\}\newline
- The ``answer'' field must be either: AGN or Galaxy\newline
- The ``reason'' field should contain a brief explanation of your classification decision\newline
- Do not include any text outside the JSON object
\end{prompt}

\begin{prompt}
\label{prompt:task1-unguided}
\textbf{Prompt S1c: Task 1: QSO Host Galaxy Classification (Unguided).}\newline
Classify this galaxy image as either AGN or Galaxy.\newline
\textbf{Output requirements:}\newline
- Respond with a JSON object in the following format: \{``answer'': ``'', ``reason'': ``''\}\newline
- The ``answer'' field must be either: AGN or Galaxy\newline
- The ``reason'' field should contain a brief explanation of your classification decision\newline
- Do not include any text outside the JSON object
\end{prompt}

\begin{prompt}
\label{prompt:task2}
\textbf{Prompt S2: Task 2: Radio Galaxy Morphology (Guided).}\newline
\textbf{Task:} Classify the radio galaxy image from \{survey\} survey according to the Fanaroff-Riley classification scheme (Class I or Class II).\newline
\textbf{Instructions:}\newline
- FRI (Fanaroff-Riley Type I): Edge-darkened sources where the radio emission is brightest near the core and fades toward the edges. The jets are typically less collimated and more turbulent.\newline
- FRII (Fanaroff-Riley Type II): Edge-brightened sources with prominent hotspots at the outer edges of the radio lobes. The jets remain well-collimated until they reach the hotspots.\newline
\textbf{Output requirements:}\newline
- Respond with a JSON object in the following format: \{``answer'': ``'', ``reason'': ``''\}\newline
- The ``answer'' field must be either: FRI or FRII\newline
- The ``reason'' field should contain a brief explanation of your classification decision\newline
- Do not include any text outside the JSON object\newline
\newline
\textit{Note:} The \{survey\} placeholder is replaced with the corresponding survey name: FIRST for MiraBest\_F and NVSS for MiraBest\_N.
\end{prompt}

\begin{prompt}
\label{prompt:task3}
\textbf{Prompt S3: Task 3: SED Classification (Guided).}\newline
\textbf{Task:} Classify the astronomical source shown in the attached SED plot as a `Type-1 AGN', `Type-2 AGN', or `Galaxy'.\newline
\textbf{Context:}\newline
- The plot shows $\nu f_\nu$ (y-axis, log scale) vs.\ Wavelength (x-axis, log scale).\newline
- The upper x-axis shows rest-frame wavelength and the lower x-axis shows observed wavelength.\newline
- The HSC $g$, $r$, $i$, $z$, $y$ bands, the Euclid $Y$, $J$, $H$ bands, the AKARI N2, N3, N4, S7, S9W, S11, L15, L18, L24 bands, and the WISE W1, W2, W3, W4 bands are marked on the plot (if available).\newline
- \{REDSHIFT\_BLOCK\}\newline
\textbf{Instructions:}\newline
Normal galaxies are entirely driven by stellar processes, peaking with starlight in the optical/NIR and star-heated cold dust in the MIR. Type 1 AGNs outshine their host galaxies across the board, dominated by the naked accretion disk in the UV/optical and the intensely heated inner dust torus in the MIR. Type 2 AGNs have their central engines hidden behind thick dust, leaving their UV/optical to appear as normal star-dominated host galaxies, while their MIR reveals the hidden monster via the glowing, re-radiating dust torus.\newline
\{REDSHIFT\_INSTRUCTION\}\newline
\textbf{Output requirements:}\newline
- Respond with a JSON object in the following format: \{``answer'': ``'', ``reason'': ``''\}\newline
- The ``answer'' field must be either: Type-1 AGN, Type-2 AGN, or Galaxy\newline
- The ``reason'' field should contain a brief explanation of your classification decision\newline
- Do not include any text outside the JSON object\newline
\newline
\textit{Note:} The \{REDSHIFT\_BLOCK\} placeholder is replaced at runtime. If redshift is provided: ``Redshift (z) = \{z\} $\pm$ \{z\_err\}''; otherwise: ``Redshift: not provided.'' The \{REDSHIFT\_INSTRUCTION\} placeholder is similarly replaced with the corresponding guidance.
\end{prompt}

\begin{prompt}
\label{prompt:task4}
\textbf{Prompt S4: Task 4: Light Curve Classification (Physical/Guided).}\newline
\textbf{Task:} Classify this multi-band light curve showing flux over time across six photometric bands ($u$, $g$, $r$, $i$, $z$, $y$).\newline
Classify the source into one of five categories based on its variability pattern:\newline
\textbf{Instructions:}\newline
- AGN: Stochastic, red-noise variability from accretion disk fluctuations around a supermassive black hole. Continuous process with no defined periodicity, persisting over long epochs.\newline
- SNIa: Thermonuclear explosion of a white dwarf. Fast rise to peak brightness followed by a characteristic exponential decline powered by radioactive decay (weeks timescale). A one-time event.\newline
- TDE (Tidal Disruption Event): A star disrupted by a supermassive black hole. Rapid rise followed by a smooth power-law decline as stellar debris falls back. Sustained high-temperature emission.\newline
- RRL (RR Lyrae): Short-period pulsating star with a period of hours to $\sim$1 day. Strictly periodic, sawtooth-like light curve with rapid rise and slow decline.\newline
- Mira: Long-period pulsating AGB star. Periodic variability over months to years, with large amplitude and smooth sinusoidal-like variations.\newline
\textbf{Output requirements:}\newline
- Respond with a JSON object in the following format: \{``answer'': ``'', ``reason'': ``''\}\newline
- The ``answer'' field must be one of: AGN, SNIa, TDE, RRL, or Mira\newline
- The ``reason'' field should contain a brief explanation of your classification decision\newline
- Do not include any text outside the JSON object
\end{prompt}

\begin{prompt}
\label{prompt:task4-phenom}
\textbf{Prompt S4b: Task 4: Light Curve Classification (Phenomenological).}\newline
You are an expert astronomical data analyst. Classify an astronomical object strictly based on the morphological and geometric patterns in its multi-band photometric light curve. Focus on visual shape, time scales, and color evolution.\newline
Classify into exactly one of 5 categories:\newline
1.~\textbf{AGN}: Continuous, non-periodic stochastic random walk persisting months to years. All bands wander synchronously (red noise).\newline
2.~\textbf{SN Ia}: Single asymmetric pulse; fast rise ($\sim$15 to 20\,d), rapid decay to baseline. Strong chromatic divergence after peak: blue bands ($u$,$g$) drop steeply, red bands ($z$,$Y$) decline slower with a ``secondary bump'' 20 to 30\,d post-peak.\newline
3.~\textbf{TDE}: Single asymmetric pulse; gradual rise, smooth protracted decay. Achromatic: all bands decline strictly parallel (constant color), no crossing or secondary bumps.\newline
4.~\textbf{RRL}: Appears as dense scatter (white noise) over long timelines due to severe aliasing of sub-day periodicity by sparse sampling. No local smoothness.\newline
5.~\textbf{Mira}: Massive smooth arch or monotonic trend over months to years; exceptionally large amplitude dominates noise.\newline
Analyze geometric shape, peak width, smoothness vs.\ noise, and whether bands evolve parallel or diverge.\newline
\textbf{Output requirements:}\newline
- JSON object: \{``answer'': ``'', ``reason'': ``''\}. Answer must be one of: AGN, SNIa, TDE, RRL, Mira.\newline
- Do not include any text outside the JSON object.
\end{prompt}

\begin{prompt}
\label{prompt:task4-unguided}
\textbf{Prompt S4c: Task 4: Light Curve Classification (Unguided).}\newline
Classify this astronomical light curve into one of five categories: AGN, SNIa, TDE, RRL, or Mira.\newline
\textbf{Output requirements:}\newline
- Respond with a JSON object: \{``answer'': ``'', ``reason'': ``''\}\newline
- The ``answer'' field must be exactly one of: AGN, SNIa, TDE, RRL, Mira\newline
- The ``reason'' field should contain a brief explanation of your classification decision\newline
- Do not include any text outside the JSON object
\end{prompt}

\begin{prompt}
\label{prompt:task5.1}
\textbf{Prompt S5.1: Task 5: Spectral Interpretation, Q1 (Guided).}\newline
\textbf{Task:} Analyze this optical spectrum of a galaxy to determine whether the spectrum contains BOTH H$\alpha$ (rest wavelength 6563\,\AA) and H$\beta$ (rest wavelength 4861\,\AA) emission lines.\newline
\textbf{Context:}\newline
- The spectrum is shown in the observed frame\newline
- H$\alpha$ and H$\beta$ may be redshifted out of the optical wavelength range or may be weak/non-detectable\newline
- Emission lines appear as peaks above the continuum\newline
\textbf{Output requirements:}\newline
- Respond with a JSON object in the following format: \{``answer'': ``'', ``reason'': ``''\}\newline
- The ``answer'' field must be either: True or False\newline
- Answer True if both lines are present, False otherwise\newline
- The ``reason'' field should contain a brief explanation\newline
- Do not include any text outside the JSON object
\end{prompt}

\begin{prompt}
\label{prompt:task5.2}
\textbf{Prompt S5.2: Task 5: Spectral Interpretation, Q2 (Guided).}\newline
\textbf{Task:} Analyze this optical spectrum of a galaxy to determine whether this object is a Broad-Line AGN (BLAGN).\newline
\textbf{Context:}\newline
- A BLAGN is characterized by broad emission lines (FWHM $>$ 1000\,km/s), particularly in H$\alpha$\newline
- The broad line component is wider than typical narrow-line regions\newline
- Type-1 AGN / Seyfert 1 classification\newline
- The H$\alpha$ feature could be blended with [N\,\textsc{ii}] lines which may complicate the profile\newline
- Look for asymmetric or broadened emission line profiles in the Balmer lines (H$\alpha$, H$\beta$)\newline
\textbf{Output requirements:}\newline
- Respond with a JSON object in the following format: \{``answer'': ``'', ``reason'': ``''\}\newline
- The ``answer'' field must be either: True or False\newline
- Answer True if this is a BLAGN, False otherwise\newline
- The ``reason'' field should contain a brief explanation\newline
- Do not include any text outside the JSON object
\end{prompt}

\begin{prompt}
\label{prompt:task5.3}
\textbf{Prompt S5.3: Task 5: Spectral Interpretation, Q3 (Guided).}\newline
\textbf{Task:} Analyze this optical spectrum of a galaxy and classify it using the BPT (Baldwin-Phillips-Terlevich) diagnostic diagram.\newline
\textbf{Context:}\newline
The BPT diagram uses the line ratios:\newline
- $\log(\text{[N\,\textsc{ii}]}\,6584\,/\,\text{H}\alpha)$ on the x-axis\newline
- $\log(\text{[O\,\textsc{iii}]}\,5007\,/\,\text{H}\beta)$ on the y-axis\newline
The four classification regions are:\newline
1. Star-Forming: Low [N\,\textsc{ii}]/H$\alpha$ and low [O\,\textsc{iii}]/H$\beta$ (ionization dominated by young stars)\newline
2. Composite: Intermediate region between SF and AGN (mixed ionization sources)\newline
3. Seyfert: High [O\,\textsc{iii}]/H$\beta$, high [N\,\textsc{ii}]/H$\alpha$ (AGN-dominated ionization)\newline
4. LINER: High [N\,\textsc{ii}]/H$\alpha$, low [O\,\textsc{iii}]/H$\beta$ (low-ionization nuclear emission region)\newline
\textbf{Output requirements:}\newline
- Respond with a JSON object in the following format: \{``answer'': ``'', ``reason'': ``''\}\newline
- The ``answer'' field must be one of: Star-Forming, Composite, Seyfert, or LINER\newline
- The ``reason'' field should contain a brief explanation\newline
- Do not include any text outside the JSON object
\end{prompt}


\end{document}